\documentclass[10pt,journal,cspaper,compsoc]{IEEEtran}
\usepackage[pdftex]{graphicx}
\usepackage[cmex10]{amsmath}
\usepackage{algorithmic}
\usepackage{array}
\usepackage{mdwmath}
\usepackage{mdwtab}
\usepackage[ruled,vlined]{algorithm2e}
\usepackage{bigstrut}
\usepackage{bm}
\usepackage{cite}
\usepackage{amsmath}
\usepackage{amssymb}
\usepackage{multirow}
\usepackage{url}
\usepackage{hyperref}
\usepackage{xcolor}

\newcommand{\argmin}[1]{\underset{#1}{\operatorname{arg}\,\operatorname{min}}\;}
\newcommand\T{{\hspace{-1pt}\intercal}}

\def\figvspace{{\vspace{-4mm}}}
\def\eqnvspace{{\vspace{-5mm}}}

\begin{document}

\title{Joint Multi-Leaf Segmentation, Alignment, and Tracking for Fluorescence Plant Videos}

\author{Xi Yin, Xiaoming Liu, Jin Chen, David M. Kramer}

\IEEEcompsoctitleabstractindextext{
\begin{abstract}

This paper proposes a novel framework for fluorescence plant video processing.
The plant research community is interested in the leaf-level photosynthetic analysis within a plant.
A prerequisite for such analysis is to segment all leaves, estimate their structures, and track them over time. 
We identify this as a joint multi-leaf {\bf segmentation, alignment,} and {\bf tracking} problem.
First, leaf segmentation and alignment are applied on the last frame of a plant video to find a number of well-aligned leaf candidates.
Second, leaf tracking is applied on the remaining frames with leaf candidate transformation from the previous frame.
We form two optimization problems with shared terms in their objective functions for leaf alignment and tracking respectively. 
A quantitative evaluation framework is formulated to evaluate the performance of our algorithm with four metrics.
Two models are learned to predict the alignment accuracy and detect tracking failure respectively in order to provide guidance for subsequent plant biology analysis. 
The limitation of our algorithm is also studied. 
Experimental results show the effectiveness, efficiency, and robustness of the proposed method.

\end{abstract}

}

\maketitle

\IEEEdisplaynotcompsoctitleabstractindextext

\IEEEpeerreviewmaketitle

\section{Introduction}
\begin{figure}[t]
\begin{center}
\includegraphics[width=.48\textwidth]{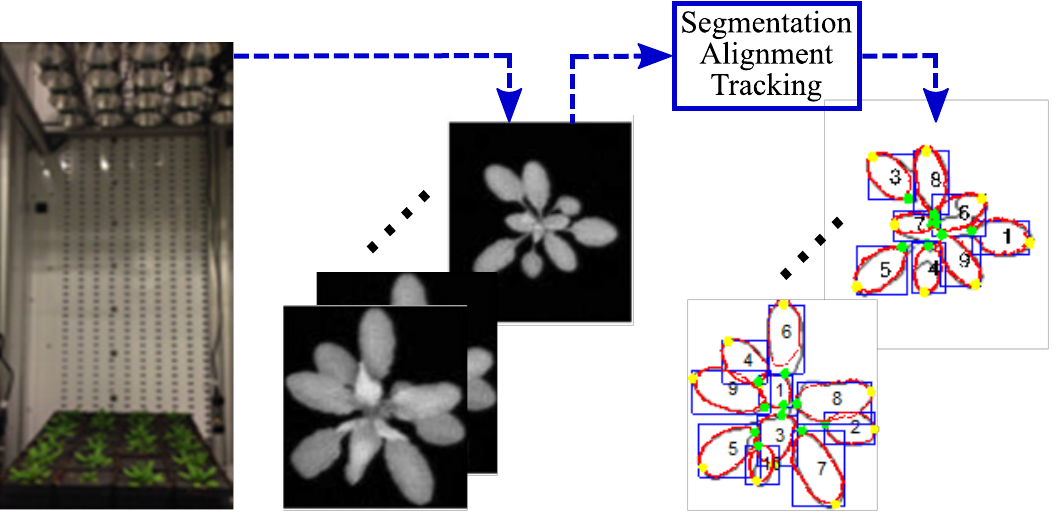}
\end{center} 
\vspace{-3mm}
\caption{{\small 
Given a fluorescence plant video captured during its growth period, our algorithm performs multi-leaf SAT jointly, \textit{i.e.}, estimating unique and consistent-over-time labels for all leaves and their individual leaf structure like leaf tips.}} 
\label{fig:concept}
\figvspace
\end{figure}

\IEEEPARstart{P}{lant} phenotyping~\cite{fiorani2013future} refers to a set of methodologies and protocols used to measure plant growth~\cite{jansen2009simultaneous}, architecture~\cite{trachsel2011shovelomics}, composition~\cite{wilson2004dissection}, and etc.
In contrast to the manual observation-based methods, the automatic image-based approaches for plant phenotyping have gained more attention recently~\cite{scharr2014annotated, hartmann2011htpheno}.
As shown in Figure~\ref{fig:concept}, plant researchers conduct large-scale experiments in a chamber with controlled temperature and lighting conditions.
Ceiling-mounted fluorescence cameras capture images of a plant during its growth period~\cite{nedbal2004chlorophyll}.
The pixel intensities of the image indicate the {\it photosynthetic efficiency} (PE) of the plant.
Given such a high-throughput imaging system, the massive data calls for advanced visual analysis in order to study a wide range of plant physiological problems~\cite{zhang2012natural}, e.g., the heterogeneity of PE among the leaves. 
Therefore, the {\it leaf-level} visual analysis is fundamental to automatic plant phenotyping.

This paper focuses on the processing of the rosette plants where the leaves are at a similar height and form a circular arrangement.
Our experiments are mainly conducted on {\it{Arabidopsis thaliana}}, which is the first plant to have its genome sequenced~\cite{arabidopsis2000analysis}.
Due to its rapid life cycle, prolific seed production, and easiness to cultivate in the restricted space, Arabidopsis is the most popular and important model plant~\cite{TAIR} in the plant research community. 
An automatic image analysis method for Arabidopsis, which is the main focus of this paper, is of essential importance for high-throughput plant phenotyping studies. 
Given a fluorescence plant video, our method performs multi-leaf Segmentation, Alignment, and Tracking (SAT) {\it jointly}.
Specifically, leaf segmentation~\cite{teng2009leaf} segments each leaf from the plant.
Leaf alignment~\cite{yin2014a} estimates the leaf structure.
Leaf tracking~\cite{Vylder2011} associates the leaves over time.
This multi-leaf analysis is a challenging problem due to several factors.
First, the image resolution is low where the small leaves are even hard to be recognized by humans.
Second, there are various degrees of overlap among leaves, which make it difficult to segment each leaf boundary.
Third, leaves within a plant exhibit various shapes, sizes, and orientations, which also change over time.
Therefore, an effective algorithm should be developed to handle all these challenges.

To the best of our knowledge, there is no previous work focusing on leaf SAT simultaneously from plant videos.
To solve this new problem, we develop two optimization algorithms.
Specifically, leaf segmentation and alignment are based on Chamfer Matching (CM)~\cite{barrow1977parametric}, which is a well-known algorithm to align one object in an image with a given template.
However, classical CM does not work well for aligning multiple overlapping leaves.
Motivated by crowd segmentation~\cite{leibe2005pedestrian}, where the number and locations of the pedestrians are estimated simultaneously, we propose a novel framework to jointly align multiple leaves in an image.
First we generate a large set of leaf templates with various shapes, sizes, and orientations.
Applying all templates to the edge map of a plant image leads to the same amount of transformed leaf templates.
We adopt the local search method for optimization to select a subset of leaf candidates that can best explain the edge map of the test image.

While leaf segmentation and alignment work well for one image, applying it to every video frame independently does not enable tracking - associating aligned leaves over time.
Therefore, we formulate leaf tracking on one frame as a problem of transforming multiple aligned leaf candidates from the previous frame.
The tracking optimization initialized with results of the previous frame can converge very fast and thus results in enhanced leaf association and computational efficiency.

In order to estimate the alignment and tracking accuracy, two quality prediction models are learned respectively.
We develop a quantitative analysis with four metrics to evaluate the multi-leaf SAT performance.
Furthermore, the limitation of our algorithm is studied. 
In summary, we make four contributions:

$\diamond$ We identify a new computer vision problem of joint multi-leaf SAT from plant videos. We collect a dataset of {\it{Arabidopsis}} and make it publicly available.

$\diamond$ We propose two optimization algorithms to solve this multi-leaf SAT problem.

$\diamond$ We develop two quality prediction models to predict the alignment accuracy and tracking failure. 

$\diamond$ We set up a quantitative evaluation framework to jointly evaluate the performance.

Compared to our earlier work~\cite{yin2014a, yin2014b}, we have made five main changes:
$1)$ One term is modified in the tracking objective function.
The proposed method is superior to~\cite{yin2014a, yin2014b} on a larger dataset.
$2)$ We develop two quality prediction models. 
$3)$ We enhance the performance evaluation procedure and add one metric to evaluate segmentation accuracy.
$4)$ We study the limitation of our tracking algorithm and show its robustness to leaf template transformation.
$5)$ We extend our method to apply on RGB images~\cite{scharr2014annotated} and compare the segmentation results to~\cite{pape20143}.

\begin{figure*}[t]
\begin{center}
\includegraphics[trim=0 0 0 0, clip, width=.96\textwidth]{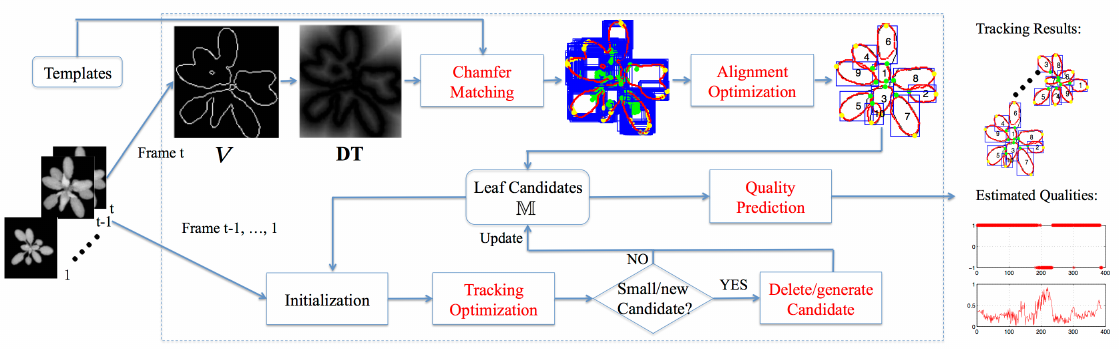}
\end{center}
\caption{\small Overview of the proposed joint multi-leaf SAT. Given a plant video with $t$ frames, the proposed method outputs the SAT results on each frame, and two prediction curves on the quality of alignment and tracking for each leaf.}
\label{fig:overview}
\figvspace
\end{figure*}

\section{Prior Work}
Plant image analysis has been studied in computer graphics~\cite{quan2006image, bradley2013image, li2013analyzing} and computer vision~\cite{chene2012use,teng2009leaf}. 
For example, a leaf shape and appearance model is proposed to render photo-realistic images of a plant~\cite{quan2006image}.
A data-driven leaf synthesis approach is developed to produce realistic reconstructions of dense foliage~\cite{bradley2013image}.
These models may not be applied to fluorescence images due to the lack of leaf appearance information.
There are prior computer vision work on tasks such as leaf segmentation~\cite{chene2012use,teng2009leaf}, alignment~\cite{cerutti2013understanding,yin2014a}, tracking~\cite{Vylder2011, yin2014b}, and identification~\cite{Mouine2012, kumar2012leafsnap, cerutti2013model}.
However, most previous studies focus on only one or two of these tasks.
In contrast, our method addresses three tasks of leaf SAT.

\noindent{\bf{Leaf Segmentation}} 
can be classified into two categories: $1)$ segmentation of a detached leaf from natural~\cite{cerutti2011parametric,chen2012image,wang2008classification,li2012leaf,cerutti2013understanding} or clean background~\cite{kumar2012leafsnap}; $2)$ pixel-wise segmentation of each leaf from a plant~\cite{cerutti2013model,pape20143}.
Methods in the first category are usually used as the first step for leaf classification or species identification.
~\cite{kumar2012leafsnap} uses pixel-based color classification for leaf segmentation from a white background.
~\cite{cerutti2013model} proposes active contour deformation method for compound leaf segmentation and identification. 

Our work belongs to the second category. 
It is very challenging due to leaf variation and overlapping.
Tsaftaris et al. organized a collation study of leaf segmentation on rosette plants in 2015~\cite{2014LSC, scharr2015leaf}. 
Our method is evaluated with other three methods. 
Two of them are based on superpixels and watershed transformation segmentation. 
~\cite{pape20143} uses distance map-based leaf center detection and leaf split points detection for leaf segmentation.

\noindent{\bf{Leaf Alignment}} aims to find the structure of a leaf, which is useful for leaf segmentation.
~\cite{cerutti2011parametric} deforms a polygonal model to leaf shape fitting, where the base and tip points are used to define a leaf template.
The same points are used in~\cite{cerutti2013understanding} to model leaf shapes and deform templates.
Similarly, we use these two points on our leaf templates for alignment.
Our novelty lies in solving leaf segmentation and alignment jointly by extending CM to align multiple potentially overlapping leaves in an image.

\noindent{\bf{Leaf Tracking}} models leaf transformation over time.
A probabilistic parametric active contour model is applied for leaf segmentation and tracking to automatically measure the temperature of leaves in~\cite{Vylder2011}.
However leaves on those images are well separated without any overlap and the active contours are initialized via the ground truth segments, which is hard to achieve in real-world applications. 
~\cite{aksoy2015modeling} segments all leaves in a video separately and employs a merging procedure to group the segments by exploiting the angle properties of the leaves.
~\cite{dellen2015growth} proposes a graph-based tracking algorithm by linking leaf detections across neighboring frames.
All of them treat tracking as a post processing {\it after} leaf segmentation on individual frame.
In contrast, we employ a leaf template transformation to transfer the segmentation and alignment results between continuous frames.

\section{Our Method}
Figure~\ref{fig:overview} shows our framework. 
Given a plant video, we first apply leaf segmentation and alignment on the last frame to generate a number of well-aligned leaf candidates.
Leaf tracking is considered as an alignment problem with the leaf candidates initialized from a previous frame.
During tracking, a leaf candidate whose size is smaller than a threshold is deleted.
A new candidate is detected and added for tracking when there is a certain region of the image mask that is not covered by the existing leaf candidates.
Two prediction models are learned to investigate the alignment and tracking quality respectively.
All notations are summarized in Table~\ref{tab:notation}.

\begin{table}[t!]
\caption{\small Notations.}
\scriptsize
\centering
\begin{tabular}{@{}c| l@{}}
\hline
Notation & Definition \\ \hline
$\bm{V},\bm{m}$ & the edge map and mask of a test image\\ \hline
$\bm{U},\bm{M}$ & the edge map and mask of a leaf template\\ \hline
$\bm{\tilde{U}}, \bm{\tilde{M}}$ & the edge map and mask of a transformed leaf template\\ \hline
$\bf{DT}$ & the distance transform image of $\bm{V}$\\ \hline
$H,S,R$ & the numbers of template shapes, sizes, and orientations\\ \hline
$N$ & the total number of leaf templates, $N=HSR$\\ \hline
$N_1$ & the number of transformed leaf templates for optimization\\ \hline
$J, G$ & the objective functions for alignment and tracking\\ \hline
$a$ & the diagonal length of $\bm{V}$ \\ \hline
$(c^x, c^y)$ & the center of a plant image\\ \hline
$(c_n^x, c_n^y)$ & the center of the ${n}^{th}$ leaf candidate\\ \hline
$\mathbb{L}, \mathbb{L}_1$ & the sets of $N$ and $N_1$ transformed templates\\ \hline
$\bm{A}$ & a $N_1 \times K$ matrix collecting all $\bm{\tilde{M}}_n$ from $\mathbb{L}_1$\\ \hline
$K$ & the number of pixels in the test image\\ \hline
$\bm{x}$ & a $N_1$-dim $0$-$1$ indicator vector \\ \hline
$\bm{d}$ & a $N_1$-dim vector of CM distances in $\mathbb{L}_1$\\ \hline
$C$ & a constant value used in $J_3$\\ \hline
$D$ & the number of maximum iterations in tracking\\ \hline
$N^e$, $N^l$ & the number of estimated and labeled leaves in a frame\\ \hline
$\mathbb{M}$ & the collection of $N^e$ selected leaf candidates\\ \hline
\multirow{2}{*}{$\bm{P}$} & a set of transformation parameters $P=\{{\bm{p_n}}\}_{n=1}^{N^e}$ \\
& ${\bm{p_n}} = [\theta, r, t_x, t_y]^\T$ is the parameter for $\bm{U}_n$ \\ \hline
$\hat{\bm{t}}_{1, 2},\bm{t}_{1, 2}$ & the estimated and labeled tips for one leaf\\ \hline
$\hat{\bm{T}}, \bm{T}$ & the estimated and labeled tips for one frame \\ \hline
$\mathbb{\hat{\bm{T}}}, \mathbb{\bm{T}}$ & the collections of estimated and labeled tips for all videos\\ \hline
\multirow{2}{*}{$\mathbb{\hat{\bm{B}}}, \mathbb{\bm{B}}$} & the collections of estimated and groundtruth segmentation\\ & masks for all videos \\ \hline
$N^b$ & the total number of labeled leaves \\ \hline
$\bm{e_{la}}$ & the tip-based error normalized by the leaf length\\ \hline
$\bm{ID}$ & a $N^e\times N^l$ matrix of leaf correspondence\\ \hline
$\bm{ER}$ & a $N^e\times N^l$ matrix of tip-based errors in one frame\\ \hline
$\mathbb{ER}$ & the collection of all $\bm{ER}$ for labeled frames\\ \hline
$f$ & the number of leaf without correspondence\\ \hline
$\bm{e}_1, \bm{e}_2$ & the tip-based errors used in Algorithm~\ref{alg_performance}\\ \hline
$\tau$ & a threshold for comparing with tip-based errors \\ \hline
$F,E,T$ & the performance metrics\\ \hline
$Q_a, Q_t$ & the quality to predict alignment and tracking \\ \hline
$\bm{x}_a, \bm{x}_t$ & the features to learn quality prediction models\\ \hline
$\lambda_{1,2},\mu_{1,2}$ & the weights used in $J$ and $G$\\ \hline
$\alpha_1, \alpha_2$ & the step sizes in the gradient descent of $J$ and $G$\\ \hline
$s$ & the smallest leaf size we use \\ \hline
\end{tabular}
\label{tab:notation}
\figvspace
\end{table}

\subsection{Multi-Leaf Segmentation and Alignment}
\label{alignment}
Our segmentation and alignment algorithm consists of two steps.
First, a pre-defined set of leaf templates is applied to the edge map of a test image to generate an over-complete set of transformed leaf templates.
Second, we formulate an optimization process to select an optimal subset of leaf candidates.

\subsubsection{Candidate nomination via Chamfer matching}
Chamfer Matching (CM)~\cite{barrow1977parametric} is a well-known method used to find the best alignment between two edge maps.
Let $\bm{V}=\{\bm{v}_i\}$ and $\bm{U} =\{\bm{u}_i\} $ be the edge maps of a test image and a template respectively.
CM distance is computed as the average distance of each edge point in $\bm{U}$ with its nearest edge point in $\bm{V}$:

\begin{equation}
d(\bm{U},\bm{V}) = \frac{1}{|\bm{U}|} \sum_{\bm{u}_i \in \bm{U}} \min_{\bm{v}_j \in \bm{V}} \| \bm{u}_i - \bm{v}_j \|_2,
\label{eqn:cm}
\end{equation}
where $|\bm{U}|$ is the number of edge points in $\bm{U}$.
CM distance can be computed efficiently via a pre-computed distance transform image ${\bf{DT}}(\bm{g})=\min_{\bm{v}_j \in \bm{V}} \|\bm{g} - \bm{v}_j\|_2$, which calculates the distance of each coordinate $\bm{g}$ to its nearest edge point in $\bm{V}$.
During the CM process, an edge template $\bm{U}$ is superimposed on ${\bf{DT}}$ and the average value sampled by the template edge points $\bm{u}_i$ equals to the CM distance, \textit{i.e.}, $d(\bm{U},\bm{V}) = \frac{1}{|\bm{U}|} \sum_{\bm{u}_i \in \bm{U}} {\bf{DT}}(\bm{u}_i)$.

Given a fluorescence plant image, it is first transformed to a binary image $\bm{m}$ by applying a threshold.
The \textit{Sobel} edge detector is applied to $\bm{m}$ to generate an edge map $\bm{V}$.
The goal of leaf alignment is to transform the $2$D edge coordinates of a template $\bm{U}$ in the leaf template space to a new set of $2$D coordinates $\bm{\tilde{U}}$ in the test image space so that the CM distance is small, \textit{i.e.}, the leaf template is well aligned with $\bm{V}$.

\begin{figure}[t!]
\begin{center}
\includegraphics[trim=45 202 35 123, clip, width=.46\textwidth]{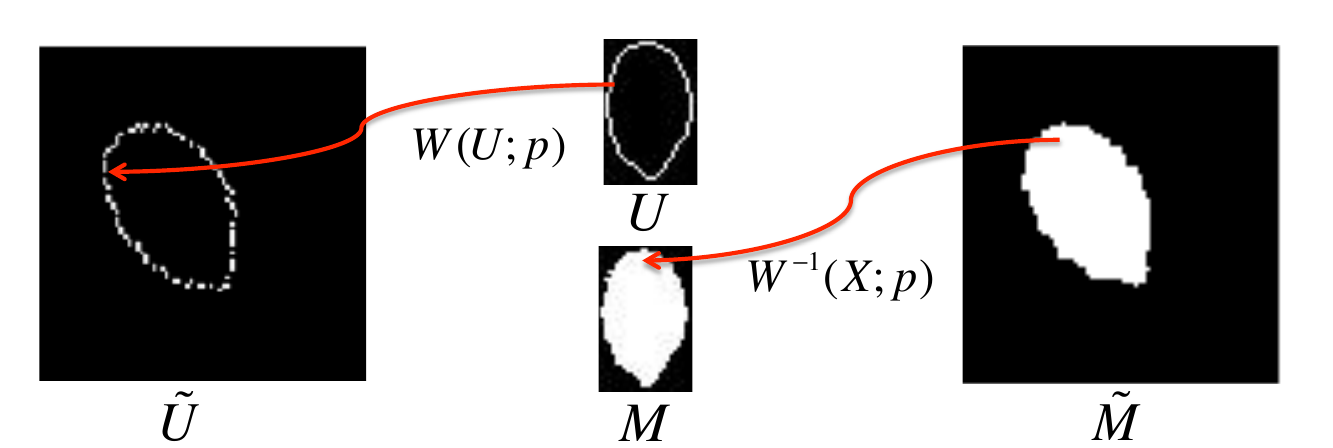}
\end{center}
\vspace{-2mm}
\caption{\small Forward and backward warping.}
\label{fig:warp}
\figvspace
\end{figure}

\noindent{\bf{Image Warping:}} In our framework, there are two types of transformations involved including forward and backward warping.
We use affine transformation that consists of scaling, rotation, and translation. 

As shown in Figure~\ref{fig:warp}, let $\bm{W}:\bm{U}\mapsto\bm{\tilde{U}}$ be a \textit{forward warping} function that transfers the 2D edge points from the template space to the test image space, parameterized by ${\bm{p}} = [\theta, r, t_x, t_y]^\T$:
\eqnvspace
\begin{equation}
\bm{\tilde{U}} =\bm{W}(\bm{U};{\bm{p}})=
r
\begin{bmatrix}
\cos \theta & -\sin\theta \\ \sin\theta & \cos\theta
\end{bmatrix}
(\bm{U} - \bar{\bm{U}}) +
\begin{bmatrix}
\bm{t_x} \\ \bm{t_y}
\end{bmatrix}
+ \bar{\bm{U}},
\label{equ:warp}
\end{equation}
where $\theta$ is the in-plane rotation angle, $r$ is the scaling factor, $t_x$ and $t_y$ are the translations along $x$ and $y$ axis respectively.
$\bar{\bm{U}}$ is the center of the leaf, \textit{i.e.}, the average of all coordinates of $\bm{U}$, which is used to model the leaf scaling and rotation w.r.t. the leaf center. 

Let $\bm{W}^{-1}:\bm{\tilde{U}}\mapsto\bm{U}$ be the \textit{backward warping} from the image space to the template space.
We denote $\bm{X}$ as a $K\times2$ matrix including all coordinates in the test image space.
Thus, $\bm{W}^{-1}(\bm{X};\bm{p})$ are the corresponding coordinates of $\bm{X}$ in the template space.
The purpose for this backward warping is to generate a $K$-dim vector $\bm{\tilde{M}} = \bm{M}(\bm{W}^{-1}(\bm{X};\bm{p}))$, which is the warped version of the original template mask ${\bm{M}}$.

\begin{figure}[t!]
\begin{center}
\includegraphics[trim=10 115 5 -14, clip, width=.42\textwidth]{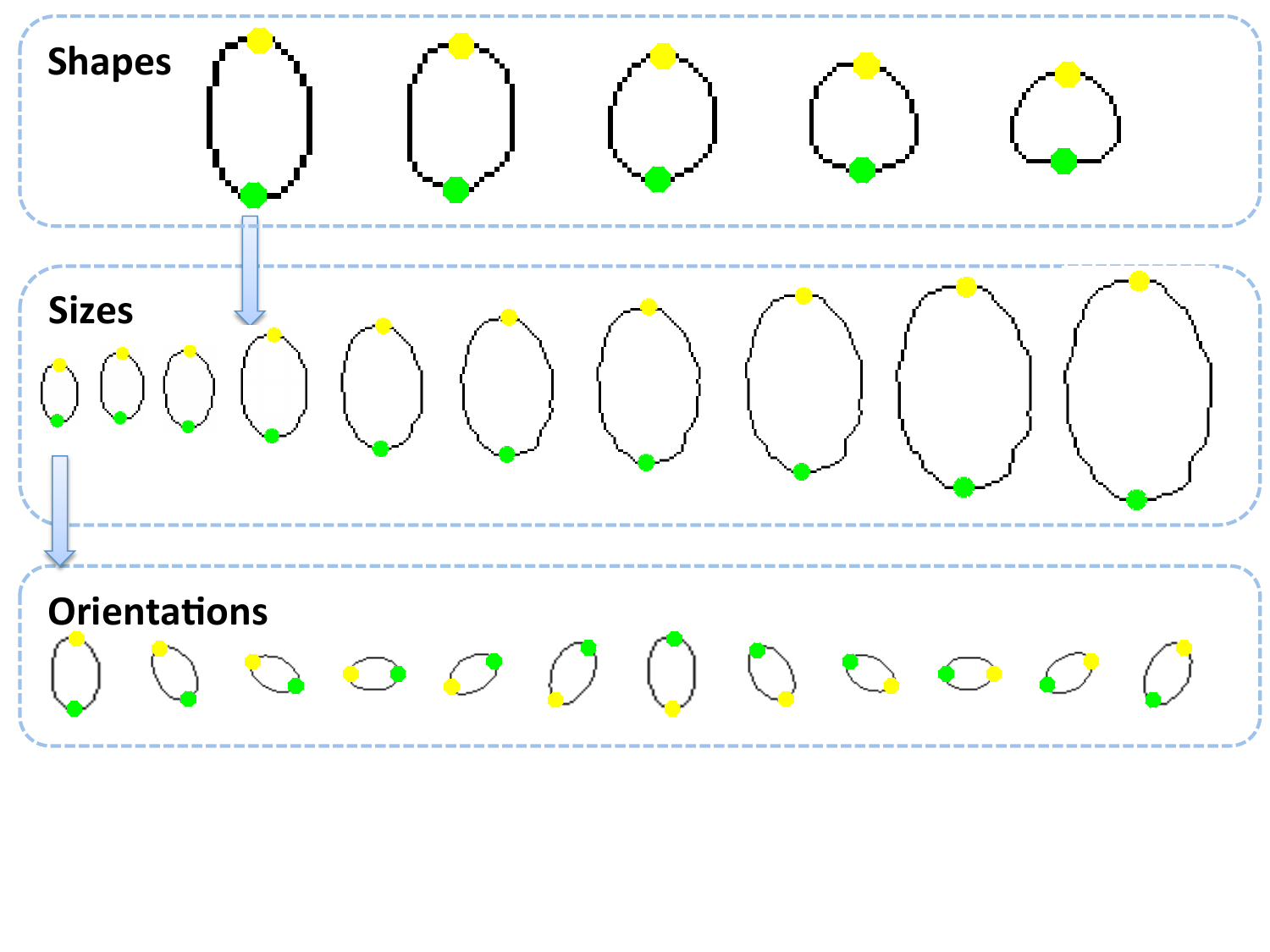}
\end{center}
\vspace{-2mm}
\caption{\small Leaf template scaling and rotation from basic template shapes. The tip labels are shown in yellow and green.}
\label{fig:template}
\figvspace
\end{figure}

\noindent{\bf{Leaf Templates:}}
Since there is a large variation in leaf shapes, it is infeasible to match leaf with one template.
We manually select $H$ basic templates (the $1^{st}$ row in Figure~\ref{fig:template}) with representative leaf shapes from the plant videos and compute their individual edge map $\bm{U}$ and mask ${\bm{M}}$.
We synthesize an over-complete set of transformed templates by selecting a discrete set of $\theta$ and $r$, which are expected to cover all potential leaf configurations in $\bm{V}$.
This leads to an array of $N=HSR$ leaf templates where $S$ and $R$ are the numbers of leaf scales and orientations respectively (Figure~\ref{fig:template}).
The yellow and green points in Figure~\ref{fig:template} are the two labeled leaf tips ${\bm{t}}$, which are used to find the corresponding leaf tips $\hat{\bm{t}}$ in $\bm{V}$ via Equation~\ref{eqn:cm}.

For each template $\bm{U}_n$, it scans through all possible locations on $\bm{V}$ and the location with the minimum CM distance is selected, which provides $t_x$ and $t_y$ optimal to $\bm{U}_n$.
Therefore, with the manually selected $\theta$, $r$, and exhaustively chosen $t_x$ and $t_y$, $N$ transformed templates are generated from $H$ basic templates.
For each transformed template, we record the 2D edge coordinates of its basic template, warped template mask, transformation parameters, CM distance and the estimated leaf tips as $\mathbb{L} = \{ \bm{U}_n, \bm{\tilde{M}}_n, {\bm{p}}_n, d_n, \hat{\bm{t}}_n\}_{n=1}^N$.
Note that $\mathbb{L}$ is an over-completed set of transformed leaf templates including the true leaf candidates as its subset.
Hence, the critical question is how to select such a subset of candidates from $\mathbb{L}$.

\subsubsection{Objective function}
The goal of leaf segmentation and alignment is to segment each leaf and estimate the structure precisely.
If the leaf candidates are well selected, there should be no redundant or missing leaves.
Each leaf candidate should be well aligned with the edge map of the test image.
This rationality leads to a three-term objective function, 
which seeks the minimal number of leaf candidates ($J_1$) with small CM distances ($J_2$) to best cover the test image mask ($J_3$).

Each image contains around $10$ leaves while the number of potential candidates in $\mathbb{L}$ is $2,880$ in our case.
The selection space needs to be narrowed down substantially. 
To do this, we compute the CM distance and the overlap ratio of each template to the test image mask. 
We remove leaf templates whose CM distance is larger than the average of all templates or whose overlap ratio is smaller than $90\%$.
Finally, we generate a new set $\mathbb{L}_1$ with $N_1$ (a few hundreds) templates.
RANSAC~\cite{fischler1981random} is not applicable here for two reasons. 
First, it is difficult to define a model or evaluation criterion for a random subset of leaf templates. 
Second, we have more outliers than inliers, which makes it hard to select the correct set in consensus.

The objective function is defined on a $N_1$-dim indicator vector $\bm{x}$, where $x_n=1$ means that the $n^{th}$ transformed template is selected and $x_n=0$ otherwise.
Hence $\bm{x}$ uniquely specifies a combination of transformed templates from $\mathbb{L}_1$.
The first term is the number of the selected leaf candidates $J_1 = \|\bm{x}\|_1$.

We concatenate $d_n$ from $\mathbb{L}_1$ to form a $N_1$-dim vector ${\bm{d}}$.
The second term, \textit{i.e.}, the average CM distance of the selected leaf candidates, is formulated as:
\begin{equation}
J_2 = \frac{{\bm{d}^\T}{\bm{x}}}{{\|\bm{x}\|_1}}.
\end{equation}

\begin{figure}[t!]
\begin{center}
\includegraphics[trim=20 96 20 25, clip, width=.46\textwidth]{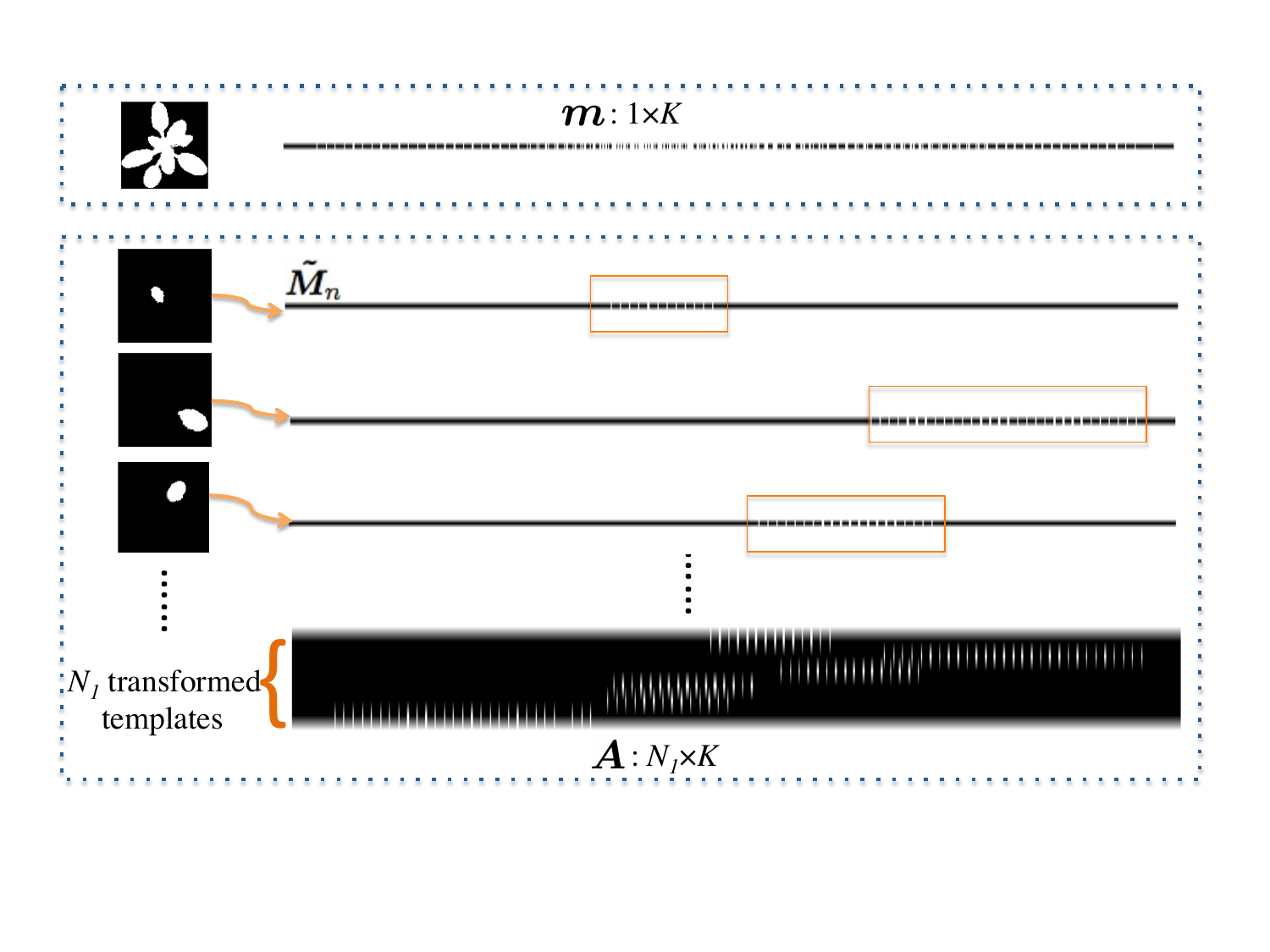}
\end{center}
\vspace{-2mm}
\caption{\small The process of generating $\bm{m}$ and $\bm{A}$.}
\label{fig:Amatrix}
\figvspace
\end{figure}

The third term is the comparison between the synthesized mask and the test image mask.
As shown in Figure~\ref{fig:Amatrix}, we convert the binary mask to a $K$-dim row vector $\bm{m}$ by raster scan.
Similarly, each warped template mask $\bm{\tilde{M}}_n$ is also a $K$-dim vector.
The collection of $\bm{\tilde{M}}_n$ from all transformed templates is denoted as a $N_1\times K$ matrix $\bm{A}$.
Note that $\bm{x}^\T\bm{A}$ is indicative of the synthesized mask except that the pixel values of the overlapping leaves are larger than $1$.
We employ the $\arctan(\cdot)$ function, similar to~\cite{Liu2008a,Liu2008b}, to convert all elements in $\bm{x}^\T\bm{A}$ to be in the range of $[0,1]$, 

\begin{equation}
f(\bm{x})= \frac {1}{\pi} \arctan(C(\bm{x}^\T\bm{A} - \frac{1}{2})) + \frac{1}{2},
\vspace{-2mm}
\end{equation}
where $C$ is a constant controlling how close $\arctan(\cdot)$ approximates the step function.
Note that the actual step function cannot be used here since it is not differentiable and thus is difficult to optimize.
The constant $\frac{1}{2}$ within the parentheses is a flip point separating where the value of $\bm{x}^\T\bm{A}$ will be pushed toward either $0$ or $1$.
Therefore, the third term becomes:

\begin{equation}
J_3= \frac{1}{K}\| f(\bm{x}) - \bm{m} \|_2^2.
\end{equation}

Finally, our objective function is:
\begin{equation}
J({\bm{x}}) = J_1 + \lambda_1 J_2 + \lambda_2 J_3,
\label{equ:alignObj}
\end{equation}
where $\lambda_1$ and $\lambda_2$ are the weights.
These three terms jointly provide guidance on what constitutes an optimal combination of leaf candidates.

\subsubsection{Local search method for optimization}
Equation~\ref{equ:alignObj} is a {\it{pseudo-Boolean function}}.
The basic algorithm~\cite{crama1990basic} is not applicable because our objective cannot be written in the required polynomial form. 
We adopt the widely used local search method to optimize Equation~\ref{equ:alignObj}. 
The local search algorithm~\cite{boros2002pseudo} for pseudo-Boolean function iteratively searches a small neighborhood of $\bm{x}$ and updates $\bm{x}$ to its neighborhood that leads to a smaller function value.

First, all elements in $\bm{x}$ are initialized as $1$, \textit{i.e.}, all transformed templates are selected.
We fix one element in $\bm{x}$ at each iteration by searching the neighborhood of $\bm{x}$ with the $n^{th}$ element being $0$ or $1$, denoted as ${\bm{x}}_{x_n=0}$ and ${\bm{x}}_{x_n=1}$.
According to the proposition $6$ in~\cite{boros2002pseudo}, a positive gradient indicates $0$ in the corresponding element of the local optimal solution. 
Therefore, each iteration, we select the element with the maximum gradient to remove redundant leaf templates. 
The gradient of the objective w.r.t. $\bm{x}$ is:
\begin{equation}
\begin{gathered}
\frac {d J}{d\bm{x}} = \mbox{sign}(\bm{x}) + \lambda_1(\frac{\bm{d}}{\|\bm{x}\|_1} -\frac{\bm{d}^\T\bm{x}}{\|\bm{x}\|_1^2} \mbox{sign}(\bm{x}) )\\
-\frac{2\lambda_{2}C}{\pi K} \bm{A}\left[ (f(\bm{x})-\bm{m}) \oslash (1 + (C(\bm{x}^\T\mathbf{A}-\frac{1}{2}))^2) \right]^\T, 
\end{gathered}
\label{equ:alignGD}
\end{equation}
where $\mbox{sign}(\bm{x})$ is a function returning the sign of each element, and $\oslash$ is the element-wise division of vectors.
In each iteration, $\bm{x}$ is updated by $\bm{x} = \bm{x} - \alpha_1 \frac {d J}{d \bm{x}}$.
The element $x_n$ with the largest gradient is chosen and fixed to be $0$ or $1$ based on the smaller value of $J({\bm{x}}_{x_n=0})$ and $J({\bm{x}}_{x_n=1})$.
Once this element is fixed, its value remains unchanged in the future iterations.
The total number of iterations is the number of transformed leaf templates $N_1$.
Finally, those elements in $\bm{x}$ equal to $1$ provide the combination of leaf candidates.

This joint leaf segmentation and alignment is applied on the last frame of a plant video to generate $N^e$ leaf candidates that are used for tracking in the remaining video frames.
We denote the set of leaf candidates selected from $\mathbb{L}_1$ as $\mathbb{M} = \{ \bm{U}_n, \tilde{\bm{M}}_n, {\bm{p}}_n, d_n, \hat{\bm{t}}_n\}_{n=1}^{N^e}$, which means the basic leaf template $\bm{U}_n$ is transformed by ${\bm{p}}_n$ to result in a leaf candidate that is well-aligned with the edge map.

\subsection{Multi-Leaf Tracking Algorithm}
Leaf tracking aims to assign the same leaf ID to the same leaf through an entire video.
In order to track all leaves over time, one way is to apply leaf segmentation and alignment framework on every frame of the video and then build leaf correspondence between consecutive frames.
However, the leaf tracking consistency is an issue due to the potentially different leaf segmentation results on different frames.
Therefore, we form an optimization problem for leaf tracking based on template transformation.

\subsubsection{Objective function}
Similar to Equation~\ref{equ:alignObj}, we formulate a three-term objective function parameterized by a set of transformation parameters $\bm{P}=\{\bm{p}_n\}_{n=1}^{N^e}$, where $\bm{p}_n$ is the transformation parameters for leaf candidate $\bm{U}_n$.

First, $\bm{P}$ is updated so that the transformed leaf candidates are well aligned with the edge map $\bm{V}$.
The first term is computed as the average CM distance of the transformed leaf candidates:
\begin{equation}
G_1 = \frac{1}{N^e}\sum_{n=1}^{N^e} d(\bm{W}(\bm{U}_n;\bm{p}_n),V).
\label{equ:G1}
\end{equation}

The second term is to encourage the synthesized mask from all transformed candidates to be similar to the test mask $\bm{m}$.
The synthesized mask of one transformed leaf candidate is ${\bf{M}}_n(\bm{W}^{-1}(\bm{X};\bm{p}_n))$, we formulate the second term as:
\begin{equation}
G_2 = \frac{1}{K} \|\sum_{n=1}^{N^e} {\bf{M}}_n(\bm{W}^{-1}(\bm{X};\bm{p}_n))-\bm{m} \|_{2}^2.
\label{equ:G2}
\vspace{-1mm}
\end{equation}

\begin{figure}[t]
\begin{center}
\includegraphics[trim=0 0 0 -5, clip, width=.45\textwidth]{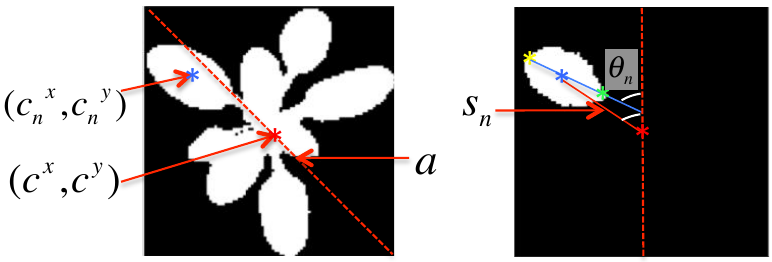}
\end{center}
\vspace{-2mm}
\caption{\small The angle difference - the long axis of leaves should point to the plant center.}
\label{fig:angle}
\figvspace
\end{figure}

One property of rosette plants such as \textit{Arabidopis} is that the long axes of most leaves point toward the center of the plant.
To take advantage of this domain-specific knowledge, the third term encourages the rotation angle $\theta$ to be similar to the direction of the leaf center to the plant center.
Figure~\ref{fig:angle} shows the geometric relation of the angle difference, which can be computed as $\frac{c_n^x- c_x}{s_n} -\sin\theta_n$, where $(c^x, c^y)$ and $(c_n^x, c_n^y)$ are the geometric centers of a plant and a leaf, \textit{i.e.}, the average coordinates of all points in $\bm{m}$ and $\bm{\tilde{U}}_n$ respectively, $s_n=\sqrt{(c_n^x- c_x)^2+(c_n^y- c_y)^2}$ is the distance between the leaf center and the plant center, and $\theta_n$ is the rotation angle.
Furthermore, since this property is more dominant for leaves far away from the plant center, we weight the above angle difference by $s_n$ and normalize it by the image size.
The third term is the average weighted angle difference:
\begin{equation}
G_3 = \frac{1}{N^e a^2}\sum_{n=1}^{N^e}\| (c_n^x- c_x) - s_n \sin\theta_n \|_2^2.
\label{equ:G3}
\end{equation}

Finally, the objective function is formulated as:
\begin{equation}
G({\bm{P}}) = G_1 + \mu_1 G_2 + \mu_2 G_3,
\label{equ:trackObj}
\end{equation}
where $\mu_1$ and $\mu_2$ are the weights.

Note the differences in two objective functions $J$ and $G$.
Since the number of leaves is fixed for tracking, $J_1$ is not needed in the formulation of $G$.
The number of leaves is relatively small during tracking.
Therefore, $\arctan(\cdot)$ is not needed since the synthesized mask is already comparable to the test image mask.

\subsubsection{Gradient descent optimization}
Given the objective function in Equation~\ref{equ:trackObj}, our goal is to minimize it by estimating $\bm{P}$, \textit{i.e.}, $\bm{P}={\bf{argmin}}_{\bm{P}} G({\bm{P}})$.
Since $G({\bm{P}})$ involves texture warping, it is a nonlinear optimization problem without a close-form solution.
We use gradient descent to solve this problem.
The derivation of $G_1$ w.r.t.~$\bm{P}$ can be written as:
\begin{equation}
\frac {d G_1}{d \bm{p}_n} = \frac{1}{N^e |\bm{U}_n|}(\triangledown{\bf{DT}}^x \frac {\partial {\bm{W}}^x}{\partial \bm{p}_n} + \triangledown{\bf{DT}}^y \frac {\partial {\bf{W}}^y}{\partial \bm{p}_n}),
\label{equ:dG1}
\end{equation}
where $\triangledown{\bf{DT}}^x$ and $\triangledown{\bf{DT}}^y$ are the gradient images of ${\bf{DT}}$ at $x$ and $y$ axis respectively.
These two gradient images only need to be computed once for each frame.
$\frac {\partial {\bm{W}}^x}{\partial \bm{p}_n}$ and $\frac {\partial {\bm{W}}^y}{\partial \bm{p}_n}$ can be easily computed from Equation~\ref{equ:warp} w.r.t.~$\theta$, $r$, $t_x$ and $t_y$ separately.

Similarly, the derivation of $G_2$ w.r.t.~$\bm{P}$ is:
\begin{equation}
\begin{gathered}
\frac {d G_2}{d \bm{p}_n} = \frac{2}{K} \Big[\sum_{n=1}^{N^e} {\bm{M}}_n(\bm{W}^{-1}(\bm{X};\bm{p}_n))-\bm{m}\Big] \cdot \\
(\triangledown{\bm{M}}_n^x \frac {\partial \bm{W}_x^{-1}}{\partial \bm{p}_n} + \triangledown {\bm{M}}_n^y \frac {\partial \bm{W}_y^{-1}}{\partial \bm{p}_n}),
\end{gathered}
\label{equ:dG2}
\end{equation}
where $\triangledown{\bm{M}}_n^x$ and $\triangledown{\bm{M}}_n^y$ are the gradient images of the template mask ${\bm{M}}_n$ at $x$ and $y$ axis respectively.
$\frac {\partial \bm{W}_x^{-1}}{\partial \bm{p}_n}$ and $\frac {\partial \bm{W}_y^{-1}}{\partial \bm{p}_n}$ can be computed based on the inverse function of Equation~\ref{equ:warp}.

The derivation of $G_3$ w.r.t.~$\theta$ is more complex than to the other three transformation parameters.
For clarity, we only present the derivative over $\theta$:
\begin{equation}
\begin{gathered}
\frac {d G_3}{d \theta_n} = \frac{2}{N^e a^2} \ [(c_n^x-c_x) - s_n \sin\theta_n] \cdot \Big[ \frac{1}{|\bm{U}_n|}\frac {\partial \bm{W}_x}{\partial \theta_n} \\
- s_n \cos\theta_n -\frac{2\sin\theta_n}{s_n|\bm{U}_n|} [(c_n^x-c_x)\frac {\partial \bm{W}_x}{\partial \theta_n} + (c_n^y-c_y)\frac {\partial \bm{W}_y}{\partial \theta_n}] \Big].
\end{gathered}
\label{equ:dG3}
\end{equation}

During optimization, $P^0$ is initialized as the transformation parameters of the leaf candidates from the previous frame and updated as $\bm{p}^t_n = \bm{p}^{t-1}_n- \alpha_2 \frac {d G}{d \bm{p}_n}$ for each leaf at iteration $t$.
Note that this is a multi-leaf \textit{joint} optimization problem because the computation of $\frac {d G_2} {d \bm{p}_n}$ involves all $N^e$ leaf candidates.
The optimization stops when $G$ does not decrease or it reaches the maximum iteration $D$.

\subsubsection{Leaf candidates update}
Given a multi-day plant video, we apply leaf segmentation and alignment algorithm on the last frame to generate $\mathbb{M}$ and employ the leaf tracking toward the first frame.
Due to plant growth and leaf occlusion, the number of leaves may vary throughout the video.
If the area of any leaf candidate at one frame is less than a threshold $s$ (defined as the number of pixels), we remove it from the leaf candidates.

On the other hand, a new leaf candidate can be detected and added to $\mathbb{M}$.
To do this, we compute the synthesized mask of all leaf candidates and subtract it from the test image mask $\bm{m}$ to generate a residue image for each frame.
Connected component analysis is applied to find components that are larger than $s$.
We then apply a subset of $N$ leaf templates to find a leaf candidate based on the edge map of the residue image.
The new candidate is assigned with an existing leaf ID if its overlap to a previous disappeared leaf is larger than a threshold. 
Otherwise it will be assigned with a new leaf ID. 
The new candidate is added into $\mathbb{M}$ and tracked in the remaining frames.

\subsection{Quality Prediction}
While many algorithms strive for perfect results, it is inevitable that unsatisfactory or failed results are obtained on the challenging samples.
It is critical for an algorithm to be aware of this situation so that future analysis does not rely on poor results. 
One approach to achieve this goal is to perform the {\it quality prediction} for the task, similar to quality estimation for fingerprint~\cite{lim2002fingerprint} and face~\cite{nasrollahi2008face}. 
The key tasks in our work include leaf alignment, estimating the two tips of a leaf, and leaf tracking, keeping leaf consistency over time.
Therefore, we learn two quality prediction models to predict the alignment accuracy and detect the tracking failure respectively.
The prediction can be used to select a subset of leaves with high {\it quality} for subsequent plant biology analysis~\cite{greenham2015trip}. 

\subsubsection{Alignment quality}
\label{AlignRegress}
Suppose $Q_a$ is the alignment accuracy of a leaf, which indicates how well the two tips are aligned.
We envision what factors may influence the estimation of the two tips.
First, the CM distance indicates how well the template and the test image are aligned.
Second, a well-aligned leaf candidate should have large overlap with the test image mask and small overlap with the neighboring leaves.
Third, the leaf area, angle, and distance to the plant center may influence the alignment result.
Therefore, we extract a $6$-dim feature vector $\bm{x}_a$ including: \emph{the CM distance} $d(\bm{W}(\bm{U}_n;\bm{p}_n),\bm{V})$, \emph{the overlap ratio with the test image mask} $\frac{1}{|\bm{m}|_1}{{\bm{\tilde{M}}}_n\odot \bm{m}}$, \emph{the overlap ratio with the other leaves} $\frac{\bm{\tilde{M}}_n \odot (\bm{m}-\bm{\tilde{M}}_n)}{|\bm{\tilde{M}}_n|_1}$, \emph{the area normalized by test image mask} $\frac{1}{|\bm{m}|_1}{|\bm{\tilde{M}}_n|_1}$, \emph{the angle difference} $|\theta_n-sin^{-1}\frac{c^x_n-c_x}{s_n}|$ and \emph{the distance to the plant center} $s_n$.
A linear regression model is learned by optimizing the following objective on $N^a$ training leaves with ground truth $Q_a$, which is proportional to the alignment error (details in Sec. $5.3.3$).
\begin{equation}
\bm{\omega}=\argmin{\bm{\omega}} \sum_{n=1}^{N^a}\parallel Q_a^n - \bm{\omega} \bm{x}_a^n \parallel,
\label{eqn:regress} 
\vspace{-1mm}
\end{equation}
where $\bm{\omega}$ is a $6$-dim weighting vector to predict the alignment accuracy of each leaf.

\subsubsection{Tracking quality}
\label{SVM}
Due to the limitation of our algorithm, it is possible that one leaf might diverge to the location of the adjacent leaves and results in tracking inconsistency.
We name it as a \emph{tracking failure}.
One example is shown in Figure~\ref{fig:performance}, where labeled leaf $1$ has been assigned two different IDs ($3$ and $4$) during tracking.
The change happens from frame $3$ to frame $2$.
The goal of tracking quality prediction is to detect the {\it moment} when tracking starts to fail.
We denote tracking quality as $Q_t$, where $Q_t=-1$ means a tracking failure of one leaf and $Q_t=1$ means tracking success.

Similar to Section~\ref{AlignRegress}, we first extract a $6$-dim feature vector $\bm{x}_a$ for one leaf.
However $\bm{x}_a$ alone can not predict the tracking failure because it does not include temporal information.
So we compare the features $\bm{x}_a$ of one frame with that of a reference frame $\hat{\bm{x}_a}$, which is $20$ frames before $\bm{x}_a$.
Since a tracking failure may result in abnormal changes in leaf area, angle, and distance to the center, we compute the leaf angle difference, leaf center distance, leaf overlap ratio between the current and the reference frame.
Finally, we form a $15$-dim feature vector denoted as $\bm{x}_t$, including $\bm{x}_a$, $\bm{x}_a-\bm{\hat{x}}_a$, \emph{the leaf angle difference} $\theta_n - \hat{\theta}_n$, \emph{the leaf center distance} $\sqrt{(c^x_n-\hat{c}^x_n)^2 + (c^y_n-\hat{c}^y_n)^2}$, and \emph{the leaf overlap ratio} $\frac{\bm{M}_n(\bm{W}^{-1}(\bm{X};\bm{p}_n)) \odot \bm{\hat{M}}_n(\bm{W}^{-1}(\bm{X};\hat{\bm{p}}_n))}{\bm{M}_n(\bm{W}^{-1}(\bm{X};\bm{p}_n))}$.
Given a training set $\Omega = \{(\bm{x}_t^n, Q_t^n)\}$ with $Q_t^n\in\{-1, 1\}$, a SVM classifier is learned as the tracking quality model.

\section{Performance Evaluation}
Leaf segmentation is to segment each leaf from the image.
Leaf alignment is to correctly estimate two tips of each leaf.
Leaf tracking is to keep the leaf ID consistent over the video.
In order to quantitatively evaluate the performance of joint multi-leaf SAT, we need to provide the ground truth of the pixel-level leaf segments in each frame, the two tips of each leaf, and the leaf IDs for all leaves in the video.

\begin{figure}[t]
\begin{center}
\includegraphics[width=.48\textwidth]{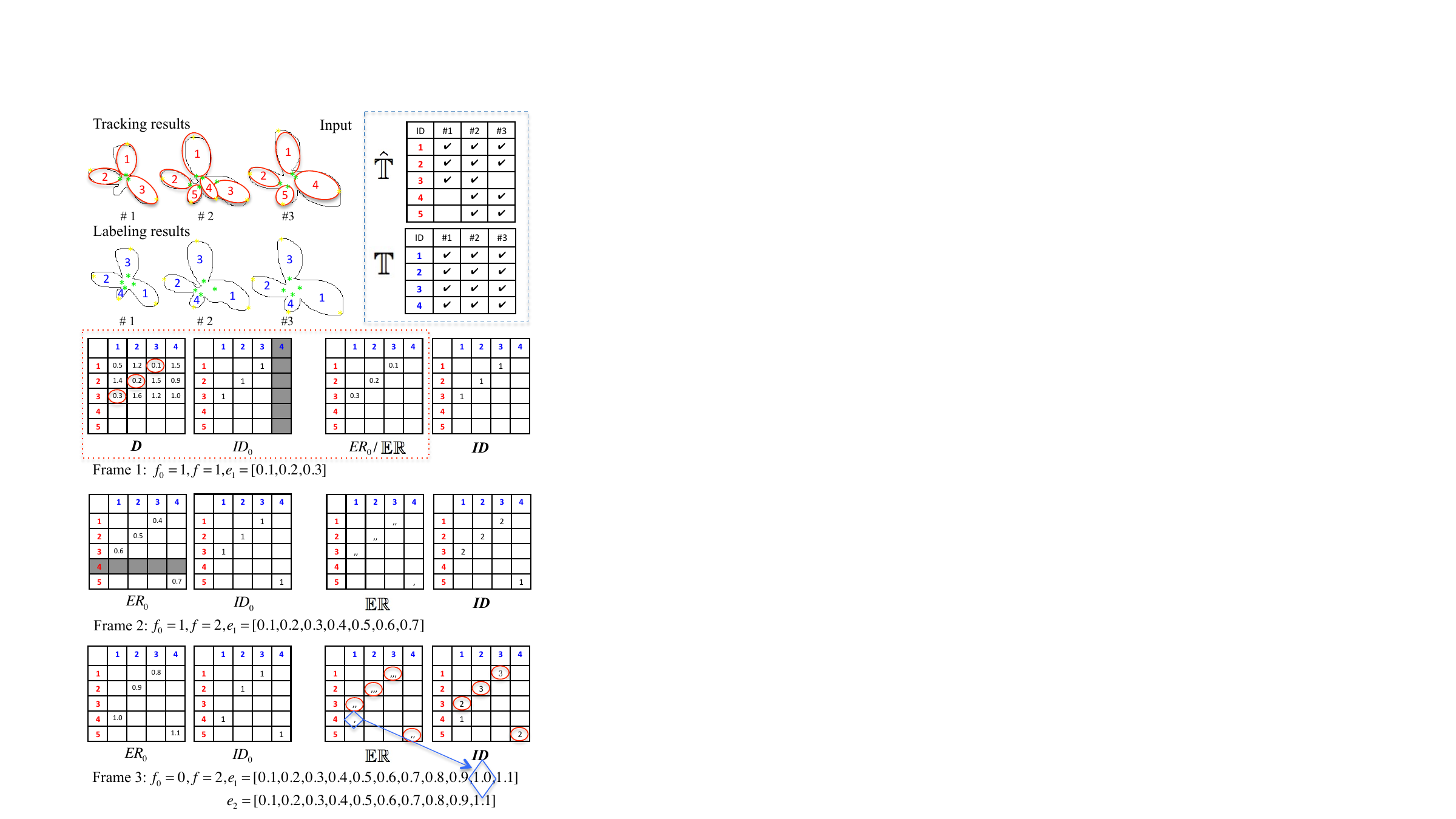}
\end{center}
\vspace{-3mm}
\caption{\small A toy example of executing Step $1$ of Algorithm~\ref{alg_performance} on one video with $3$ frames. 
In frame $1$, we illustrate the process of Algorithm~\ref{alg_match}.
Each table shows the corresponding computation of each leaf from tracking results (each row) and label results (each column).}
\vspace{-1mm}
\label{fig:performance}
\end{figure}

As shown in Figure~\ref{fig:performance}, we label the two tips of each leaf and manually assign their IDs in several frames of one video.
We record the label results in one frame as a $N^l\times 4$ matrix $\bm{T}$, where $N^l$ is the number of labeled leaves and $\bm{T}(n,:)=[\bm{t}_1^x, \bm{t}_1^y, \bm{t}_2^x, \bm{t}_2^y]$ records tip coordinates of $n^{th}$ leaf in this frame.
The collection of all labeled frames in all videos is denoted as ${\mathbb{T}}$, where $\bm{T}={\mathbb{T}}\{i,j\}(i=1,2,...,m, j=1,2,...n)$, $m$ is the number of labeled videos and $n$ is the number of labeled frames in each video.
The total number of labeled leaves in ${\mathbb{T}}$ is $N^b$.

During template transformation, the corresponding points of the transformed template tips in $\bm{V}$ become the estimated leaf tips $[\hat{\bm{t}}_1^x, \hat{\bm{t}}_1^y, \hat{\bm{t}}_2^x, \hat{\bm{t}}_2^y]$.
The leaf ID is assigned in the last frame starting from $1$ to the total number of selected leaves and kept the same during tracking.
Similar to the data structure of ${\mathbb{T}}$, the tracking results of all videos over the labeled frames is written as ${\mathbb{\bm{\hat{T}}}}$.
Given ${\mathbb{\bm{\hat{T}}}}$ and ${\mathbb{T}}$, Algorithm~\ref{alg_performance} provides our detailed performance evaluation, which is also illustrated by a synthetic example in Figure~\ref{fig:performance}.

There are two concepts involved: frame-to-frame and video-to-video correspondence.
For each estimated leaf, we need to find one corresponding leaf in the labeled frame.
Frame-to-frame correspondence aims to assign a unique leaf ID to each leaf in the frame so that the IDs are consistent with our labeled IDs.
As mentioned before, the frame-to-frame correspondence may not be consistent in the whole video due to the tracking failures or more than one leaf IDs can be assigned to the same leaf in the video.
Video-to-video correspondence aims to assign consistent and unique leaf ID to the same leaf in the entire video.

\begin{algorithm}[t]
\DontPrintSemicolon{
\renewcommand{\baselinestretch}{1.0}
\small \KwIn{Estimated leaf tips matrix $\hat{\bm{T}}$($N^e\times 4$) and labeled leaf tips matrix $\bm{T}$($N^l\times 4$).}
\KwOut{$f=|N^l-N^e|$, $\bm{ER}$, and $\bm{ID}$.}
\SetAlCapFnt{\small}
Initialize $\bm{D}$=$\bm{ER}$=$\bm{ID}$=${\bf{0}}_{N^e \times N^l}$.\\
\For{$i=1,\dots,N^e$}
{
\For{$j=1,\dots,N^l$}
{
$\bm{D}(i,j) = e_{la}(\bm{t}_i, \bm{\hat{t}}_j)$;
}
}
\For{$k=1,\dots,\mbox{min}(N^e, N^l)$}
{
$[\bm{e}_{min}, i,j] = \mbox{min}(\bm{D})$;\linebreak
${\bm{D}}(i, :) = \mbox{Inf}$;
${\bm{D}}(:, j) = \mbox{Inf}$; \linebreak
${\bm{ID}}(i, j) = 1$;
${\bm{ER}}(i, j) = \bm{e}_{min}$.
}
}
\caption{\small Build leaf correspondence $[f, \bm{ER}, \bm{ID}] = \mbox{leafMatch} (\bm{\hat{T}}, \bm{T})$.}
\label{alg_match}
\vspace{-1mm}
\end{algorithm}

\begin{algorithm}[t]
\DontPrintSemicolon{
\renewcommand{\baselinestretch}{1.0}
\small \KwIn{Tracking results $\hat{\mathbb{T}}$, label results ${\mathbb{T}}$.}
\KwOut{$F$, $E$, and $T$.}
\SetAlCapFnt{\small}
Initialize $f=0$, $\bm{e}_1=\bm{e}_2=[]$.\linebreak
1.\For{$i=1,\dots,m$}
{
$\mathbb{ER}$ = cell($N^e$, $N^l$), $\bm{ID}={\bf{0}}_{N^e \times N^l}$.\linebreak
\For{$j=1,\dots,n$}
{
$\hat{\bm{T}} =\hat{\mathbb{T}}\{i,j\}$;
$\bm{T} = \mathbb{T}\{i,j\}$;
$[f_0, \bm{ER}_0, \bm{ID}_0] = { \textbf{leafMatch}}(\hat{\bm{T}}, \bm{T})$;\linebreak
$f = f + f_0$;
$\bm{e}_1 = [\bm{e}_1, \bm{ER}_0(\bm{ER}_0\neq 0)]$;
$\mathbb{ER} = \mathbb{ER} + \bm{ER}_0$;
$\bm{ID} = \bm{ID} + \bm{ID}_0$;
}
\For{$k=1,\dots,\mbox{min}(N^e, N^l)$}
{
$[\bm{ID}_{max}, i, j] = \mbox{max}({\bm{ID}})$;\linebreak
${\bm{ID}}(i, :) = \mbox{Inf}$;
${\bm{ID}}(:, j) = \mbox{Inf}$; \linebreak
$\bm{e}_2 = [\bm{e}_2, \mathbb{ER}\{i, j\}]$;
}
}
2.\For{$\tau=0:0.01:1\;$}
{
$F(\tau) = \frac{f+\mbox{sum}(\bm{e}_1>\tau)}{N^b}$;
$E(\tau) = \mbox{mean}(\bm{e}_1\leq\tau)$;
$T(\tau) = \frac{\mbox{sum}(\bm{e}_2\leq\tau)}{N^b}$.
}
}
\caption{\small Performance evaluation process.}{} \label{alg_eva}
\label{alg_performance}
\vspace{-1mm}
\end{algorithm}

We start by building frame-to-frame leaf correspondence, as in Algorithm~\ref{alg_match} and the red dotted box in Figure~\ref{fig:performance}.
To build the leaf correspondence of $N^e$ estimated leaves with $N^l$ labeled leaves, a $N^e\times N^l$ matrix $\bm{D}$ is computed, which records all tip-based errors of each estimated leaf tips $\hat{\bm{t}}_{1,2}$ with every labeled tips $ \bm{t}_{1,2}$ normalized by the labeled leaf length:
\begin {equation}
e_{la}(\hat{\bm{t}}_{1,2}, \bm{t}_{1,2}) = \frac{||\hat{\bm{t}}_1-{\bm{t}}_1||_2 + ||\hat{\bm{t}}_2-{\bm{t}}_2||_2}{2 ||\bm{t}_1-\bm{t}_2||_2}.
\label{eqn:error}
\end{equation}

We build the leaf correspondence by finding a number of $\mbox{min}(N^e, N^l)$ minimum errors in $\bm{D}$ that do not share columns or rows, which results in $\mbox{min}(N^e, N^l)$ leaf pairs and $f=|N^l-N^e|$ leaves without correspondence.
Finally, it outputs the number of unmatched leaf $f$, $\bm{ER}$ recording tip-based errors and $\bm{ID}$ recording the leaf correspondence.
This frame-to-frame correspondence is built on all $3$ frames and the results are added into $\mathbb{ER}$ and $\bm{ID}$. 
We build the video-to-video leaf correspondence using the accumulated $\bm{ID}$.
$\bm{e}_1$ and $\bm{e}_2$ are the tip-based errors of leaf pairs with frame-to-frame and video-to-video correspondence respectively.
The difference of $\bm{e}_1$ and $\bm{e}_2$ is from estimated leaf $4$. 
While it is well aligned with labeled leaf $1$ in frame $3$, it does not have leaf correspondence in all $3$ frames together.

Finally we compute three metrics by varying a threshold $\tau$.
{\textbf{Unmatched leaf rate}} $F$ is the percentage of unmatched leaves w.r.t.~the total number of labeled leaves $N^b$.
$F$ attributes to two sources, $f$ leaves without correspondence and correspondent leaves with tip-based errors larger than $\tau$.
{\textbf{Landmark error}} $E$ is the average of all tip-based errors in $\bm{e}_1$ that are smaller than $\tau$.
{\textbf{Tracking consistency}} $T$ is the percentage of leaf pairs whose tip-based errors in $\bm{e}_2$ are smaller than $\tau$ w.r.t.~$N^b$.
These three metrics jointly estimate the accuracy in leaf counting ($F$), alignment ($E$), and tracking ($T$).

In order to quantitatively evaluate the segmentation accuracy, we annotate each image to generate a leaf segmentation mask where the pixels of the same leaf are assigned with the same number over the video. 
We add the metric \textit{``Symmetric Best Dice'' (SBD)}~\cite{scharr2014annotated} to compute the similarity between the estimated and the ground truth segmentation masks.
It is averaged across all labeled frames.
These four metrics are used to evaluate the performance of our joint framework.

\section{Experiments and Results}
\label{experiments}
\subsection{Dataset and Templates}
Our dataset includes $41$ \textit{Arabidopsis Thaliana} videos taken in a $5$-day period, which is sufficient to model the plant growth~\cite{tessmer2013functional}.
Each video has $389$ frames, with the image resolution ranging from $40\times 40$ to $100\times 100$.
For each video, we label the two tips of all visible leaves and segmentation masks of $5$ frames, each being the middle frame of a day.
In total we labeled $N^b=1,807$ leaves.
We select $10$ videos to form the training set for template generation and parameter tuning.
The remaining $31$ videos are used for testing.
The collection of all labeled tips and segmentation masks are denoted as ${\mathbb{T}}$ and ${\mathbb{B}}$.

To generate leaf templates, we select $10$ leaves with representative shapes and label the two tips for each leaf, as in Figure~\ref{fig:template}.
We select $12$ scales for each leaf shape to guarantee the scaled templates can cover all possible leaf sizes in the dataset.
For each scaled leaf template, we rotate it every $15^{\circ}$ in the $360^{\circ}$ space.
Finally, the total number of leaf templates is $N=SHR=2,880$ with $S=10, H=12, R=24$~\footnote{The dataset, labels, and templates are publicly available at: \url{http://cvlab.cse.msu.edu/project-plant-vision.html}.}.

\subsection{Experimental Setup}
For each testing video, we apply our approach and compare with four methods: {\emph{Baseline Chamfer Matching}}, {\emph{Prior Work}} ~\cite{yin2014a}, ~\cite{yin2014b}, and {\emph{Manual Results}}.

\noindent{\bf{Proposed Method}} $N$ templates are applied to the edge map of the last video frame to generate the same amount of transformed templates. 
Leaf segmentation and alignment generate $N^e$ leaf candidates for leaf tracking, which iteratively updates $\bm{P}$ according to Equation~\ref{equ:trackObj} towards the first frame.

\noindent{\bf{Baseline Chamfer Matching}} The basic idea of CM is to align one object in an image.
To align multiple leaves in a plant image, we design the \emph{baseline CM} to iteratively align one leaf at one time.
In each iteration, we apply all $N$ templates to the edge map of a test image to generate $N$ transformed leaf templates, which is the same as our first step. 
The transformed template with the minimum CM distance is selected and denoted as a leaf candidate.
We update the edge map by deleting the matched edge points of the selected leaf candidate.
The iteration continues until $90\%$ of the edge points are deleted.
We apply this method to the labeled frames of each video and build the leaf correspondence based on leaf centers.

\noindent{\textbf{Multi-leaf Alignment}}~\cite{yin2014a} 
The optimization in~\cite{yin2014a} is the same as our proposed leaf alignment on the last frame.
We apply~\cite{yin2014a} on all labeled frames and build the leaf correspondence based on leaf center distances.

\noindent{\bf{Multi-leaf Tracking}}~\cite{yin2014b} 
The difference between the proposed method and~\cite{yin2014b} includes the modified $G_3$ in Equation~\ref{equ:trackObj}. And~\cite{yin2014b} do not have the scheme to generate a new leaf candidate during tracking.

\noindent{\bf{Manual Results}} In order to find the upper bound of our proposed method, we use the ground truth labels $\mathbb{T}$ to find the optimal set of $\hat{\mathbb{T}}$.
For each labeled leaf, we find the leaf candidate with the smallest tip-based error $e_{la}$ from $N$ transformed templates.

For all methods, we record the estimated tip coordinates of all leaf candidates in the labeled frames as $\hat{\mathbb{T}}$.
The transformed template masks are used to generate an estimated segmentation mask for each frame.
We record the estimated segmentation masks of all labeled frames as $\hat{\mathbb{B}}$.
$\hat{\mathbb{T}}$ and ${\mathbb{T}}$ are used to evaluate $F$, $E$, and $T$.
$\hat{\mathbb{B}}$ and ${\mathbb{B}}$ are used to evaluate \textit{SBD}.

\begin{figure*}[t]
\begin{center}
\includegraphics[trim=0 0 0 0, clip, width=.95\textwidth]{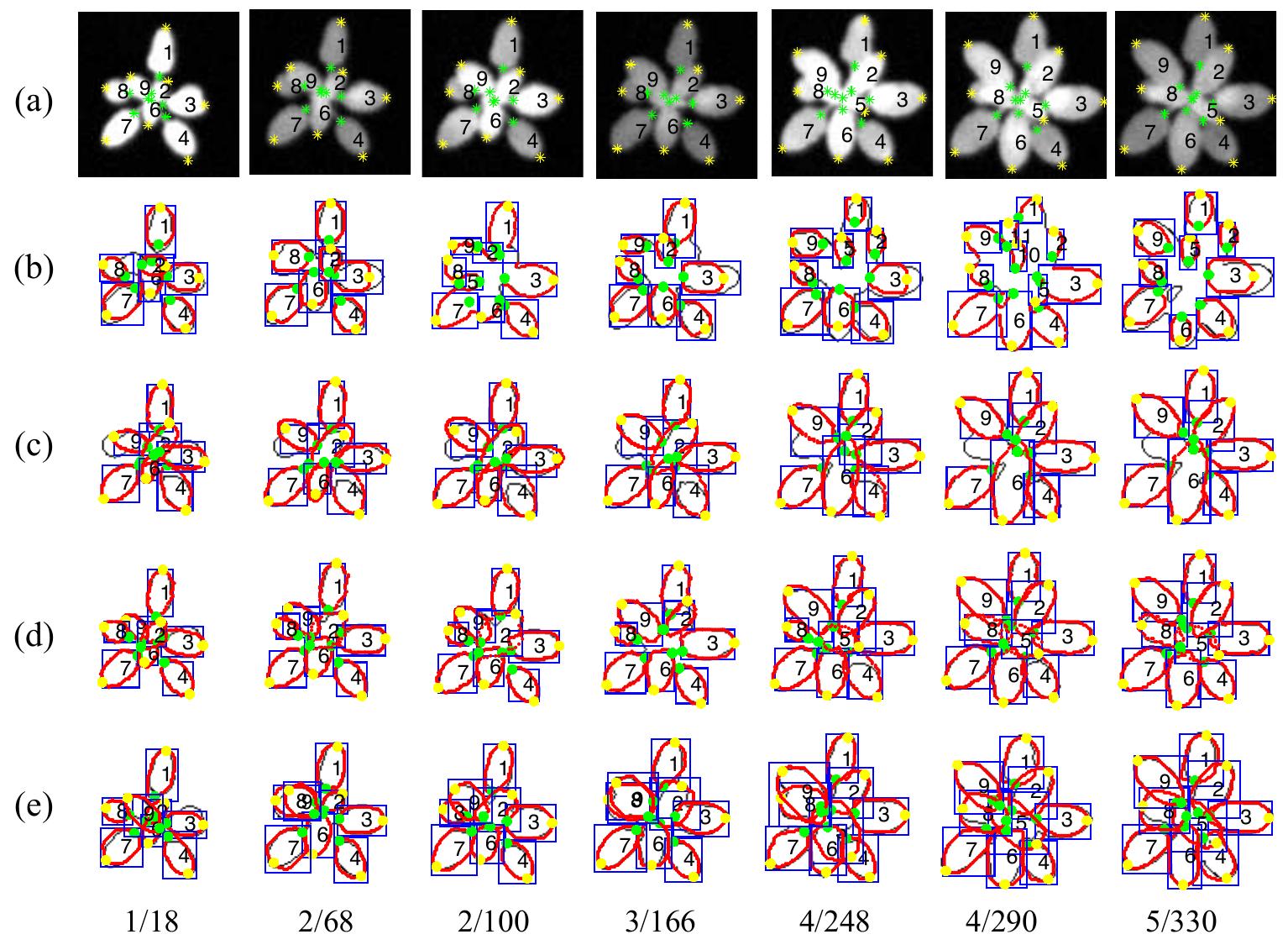}
\end{center} 
\vspace{-3mm}
\caption{\small Qualitative results: (a) ground truth labels; (b) baseline CM; (c)~\cite{yin2014b}; (d) proposed method; and (e) manual results. Each column is one frame in the video (day/frame). \textcolor{yellow}{Yellow}/\textcolor{green}{green} dots are the estimated outer/inner leaf tips. \textcolor{red}{Red} contour is $\bm{W}(\bm{U};\bm{p})$. \textcolor{blue}{Blue} box encloses the edge points matching $\bm{W}(\bm{U};\bm{p})$. The number on a leaf is the leaf ID. Best viewed in color.}
\label{fig:visual}
\end{figure*}

\begin{figure*}[t!]
\begin{center}
\includegraphics[width=.99\textwidth]{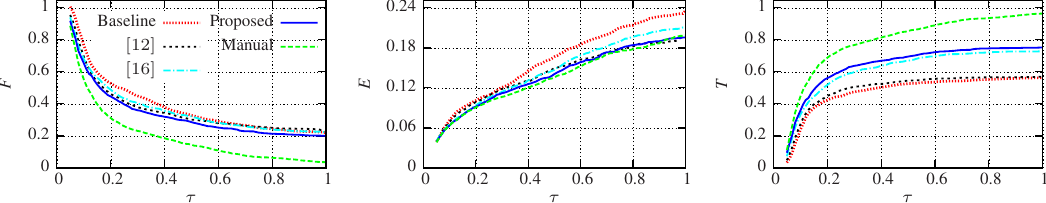}
\end{center}
\vspace{-4mm}
\caption{\small Accuracy comparison of $F$, $E$, and $T$ vs. $\tau$ for all different methods based on Algorithm~\ref{alg_performance}.}
\label{fig:curves}
\vspace{-2mm}
\end{figure*}

\subsection{Experimental Results}
\subsubsection{Performance comparison}
\noindent{\bf{Qualitative Results}} 
Figure~\ref{fig:visual} shows the results on the labeled frames of one video.
Since the baseline CM only considers CM distance to segment each leaf separately, leaf candidates are likely to be aligned around the edge points, which result in large landmark errors.
While~\cite{yin2014b} can keep the leaf ID consistent, it does not include the scheme to generate a new leaf candidate during tracking (e.g., leaf $8$ in Figure~\ref{fig:visual}).
Our proposed method performs substantially better than others.
It has the same segmentation as the labeled results and all leaves are well tracked.
Leaf $5$ is deleted when it gets too small.
Due to the limitation of a {\it finite} amount of templates, the manual results are not perfect.
However, in our tracking method, we allow template transformation under any parameters in $P$ without limiting to a finite number.

\noindent{\bf{Quantitative Results}} We first evaluate the SAT accuracy w.r.t.~$F$, $E$, and $T$. 
We set the threshold $\tau$ to vary in $[0.05:0.01:1]$ and generate the accuracy curves for all methods, as shown in Figure~\ref{fig:curves}.
When $\tau$ is small, {\it i.e.}, we have very strict requirements on the accuracy of tip estimation, all methods work well for easy-to-align leaves.
With the increase of $\tau$, more and more hard-to-align leaves with relatively large tip-based errors are considered as well-aligned leaves and contribute to the landmark error $E$ and tracking consistency $T$.
Therefore, detecting more leaves will result in higher $E$ and $T$.
It is noteworthy that our method achieves {\it lower} landmark error and {\it higher} tracking consistency while segmenting {\it more} leaves.

The baseline CM segments less leaves with higher landmark error and lower tracking consistency. 
The manual results are the upper bound of our algorithm.
Obviously $F$ will be $0$ and $T$ will be $1$ with the increase of $\tau$ because we enforce the correspondence of all labeled leaves.
But $E$ will not be $0$ due to the limitation of a finite template set.
Overall, the proposed method performs much better than the baseline CM and our prior work.
The improvement over~\cite{yin2014a} is mainly in a higher $T$,
and it improves~\cite{yin2014b} in all three metrics.
However there is still a gap between the proposed method and the manual results, which calls for future research.

The \textit{SBD}-based segmentation accuracy is shown in Table~\ref{tab:efficient}.
The proposed method is again superior to the baseline algorithm and the prior work.

\noindent{\bf{Efficiency Results}} Table~\ref{tab:efficient} shows the average execution time, which is calculated based on a Matlab implementation on a conventional computer.
Our method is superior to the baseline CM and~\cite{yin2014a}. 
It is a little slower than~\cite{yin2014b} because of the updated $G_3$ and the scheme to add leaf candidates during tracking.

\begin{table}
\caption{\small \textit{SBD} and efficiency comparison (sec./image).}
\vspace{-2mm}
\small
\centering
\begin{tabular}{c ccccc}
\hline\hline
& Baseline &~\cite{yin2014a} &~\cite{yin2014b} & Proposed & Manual\\
\hline
\textit{SBD} & 61.0 & 63.0 & 64.4 & 65.2 & 74.9\\
Time & 51.28 & 16.42 & 1.98 & 2.15 & -\\
\hline
\end{tabular}
\label{tab:efficient}
\end{table}

\begin{table}[t!]
\caption{\small Leaf segmentation \textit{SBD} accuracy ($\pm$std) comparison.}
\vspace{-2mm}
\footnotesize
\centering
\begin{tabular}{ccccc}
\hline\hline
& A1 & A2 & A3 & all\\
\hline
~\cite{pape20143} & 74.2($\pm$7.7) & 80.6($\pm$8.7) & 61.8($\pm$19.1) & 73.5($\pm$11.5)\\
Ours & 78.5($\pm$5.5) & 77.4($\pm$8.1) & 76.1($\pm$14.1) & 78.0($\pm$7.8)\\
\hline
\end{tabular}
\label{tab:segcomp}
\end{table}

\begin{figure*}[t]
\begin{center}
\includegraphics[width=.92\textwidth]{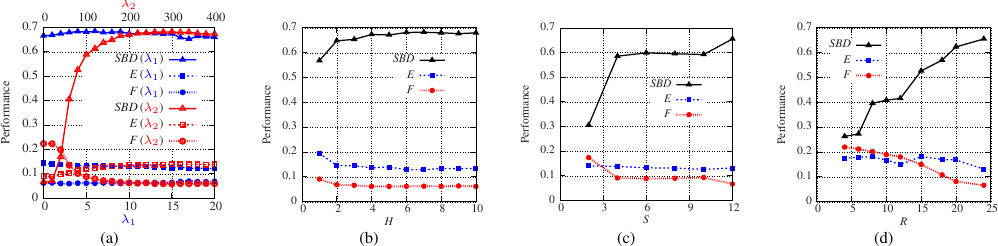}
\end{center}\vspace{-3mm}
\caption{\small Alignment parameter tuning. We show the accuracy in $E$, $F$, and \textit{SBD} when varying the coefficients of each objectve (a) and the template set size (b,c,d).}
\label{fig:alignParaTune}\figvspace
\end{figure*}

\begin{figure}[h]
\centering
\begin{tabular}{cccc}
\includegraphics[width=.1\textwidth]{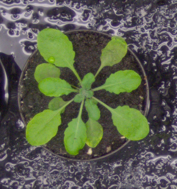} &
\includegraphics[width=.1\textwidth]{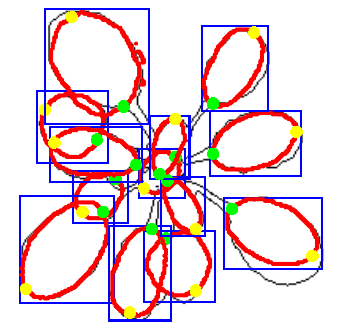} &
\includegraphics[width=.1\textwidth]{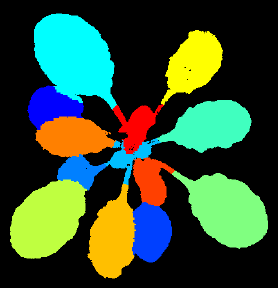} &
\includegraphics[width=.1\textwidth]{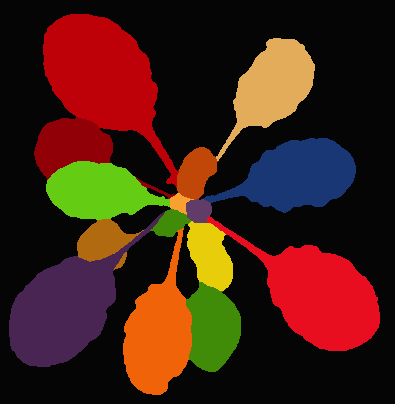} \\
\includegraphics[width=.1\textwidth]{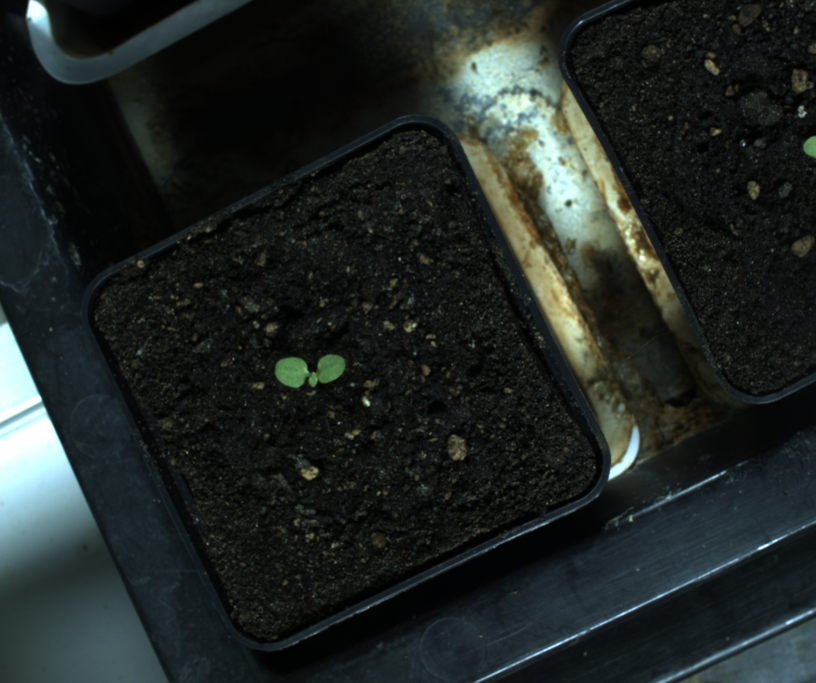} &
\includegraphics[width=.1\textwidth]{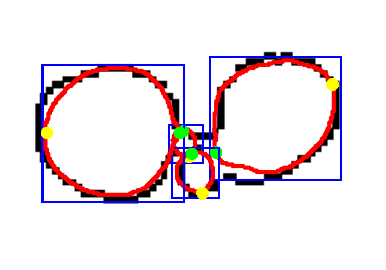} &
\includegraphics[width=.1\textwidth]{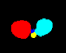} &
\includegraphics[width=.1\textwidth]{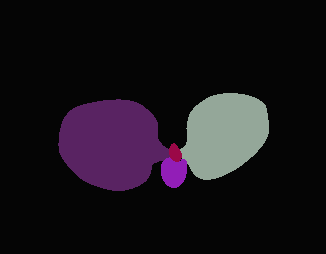} \\
(a) & (b) & (c) & (d)\\
\end{tabular}
\caption{\small Leaf segmentation results on LSC: (a) input image; (b) alignment result; (c) estimated segmentation mask; (d) ground truth segmentation mask.}
\label{fig:LSC}
\end{figure}

\noindent{\bf{Segmentation Accuracy}}
While there is no prior work focuses on the joint multi-leaf SAT, leaf segmentation has been studied especially on RGB imagery. 
For example, state-of-the-art performance~\cite{pape20143} is reported in the 2014 Leaf Segmentation Challenge (LSC)~\cite{2014LSC}.
We apply our segmentation and alignment algorithm to the LSC dataset~\cite{scharr2014annotated}, which consists of three sets of \textit{Arabidopsis} ($A1,A2$) and \textit{tobacco} ($A3$).
Two examples from $A2$ and $A3$ are shown in Figure~\ref{fig:LSC}.
Note that pre-processing of the RGB imagery is employed in order to extend our proposed method to this LSC dataset.
We compare the segmentation accuracy with~\cite{pape20143} in Table~\ref{tab:segcomp}.
Our algorithm achieves higher \textit{SBD} in A$1$, A$3$, and in average. 
The segmentation accuracy on the LSC dataset is much higher than that of our fluorescence dataset because images in~\cite{scharr2014annotated} are of higher resolution.

\subsubsection{Parameter tuning}
We explore the sensitivity of the parameters in our method. 
We use the $10$ training videos for parameter tuning in our framework.
For alignment parameter tuning, we test on all labeled frames independently and evaluate the accuracy without using tracking consistency $T$.
For tracking parameter tuning, we test on the labeled frames of each video and evaluate the accuracy using all four metrics.

Figure~\ref{fig:alignParaTune} (a) shows the alignment parameter tuning results of the weights for each objective term in Equation~\ref{equ:alignObj}. 
We first search for the optimal setting to be: $\lambda_1=4$ and $\lambda_2=300$.
We then fix one parameter and change the other and evaluate the performance at $\tau=0.4$.
We observe that $\lambda_1$ is relatively robust with some improvement from $0$ to $4$.
The performance increases tremendously as $\lambda_2$ increases, indicating that $J_3$ is crucial.
Without either term ($\lambda_1=0$ or $\lambda_2=0$), the performance is not optimal.

In order to analyze the impact of the number of leaf templates, we reduce the value in one of $H$, $S$, and $R$ at a time.
As shown in Figure~\ref{fig:alignParaTune}, the performance increases as the number of templates increases in all three parameters.
However, orientation is the most important as leaves with different orientations are more likely to have higher CM distances than leaves with different shapes or scales.

\begin{figure}[t!]
\begin{center}
\includegraphics[width=\columnwidth]{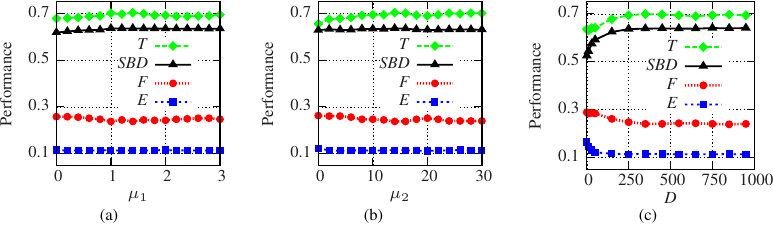}
\end{center}\vspace{-3mm}
\caption{\small Tracking parameter tuning. Accuracy w.r.t.~the coefficients in the objective and the number of iterations.}
\label{fig:trackParaTune}\figvspace
\end{figure}

Figure~\ref{fig:trackParaTune} (a,b) shows the parameter tuning results in leaf tracking framework.
Similarly, we first find the optimal weights to be: $\mu_1=1$ and $\mu_2=15$.
We fix one parameter and change the other and evaluate the performance at $\tau=0.4$.
$\mu_1$ and $\mu_2$ are relatively robust to changes.
However, they are still useful as without either term, the performance decreases.

To study the impact of the number of iterations between two frames, we change $D$ and evaluate the performance.
As shown in Figure~\ref{fig:trackParaTune} (c), the performance increases as $D$ increases.
However, it stabilizes when $D$ is larger than $300$ because the algorithm already converges before reaching the maximum iteration. 

In summary, all parameters used in our algorithm are set as: $\lambda_1=4$, $\lambda_2=300$, $C=3$, $\mu_1=1$, $\mu_2=15$, $D=300$, $\alpha_1=0.001$, $\alpha_2=0.001$, and $s=40$.

\subsubsection{Quality prediction}
\begin{figure}[t!]
\begin{center}
\includegraphics[trim=98 280 150 92, clip, width=.5\textwidth]{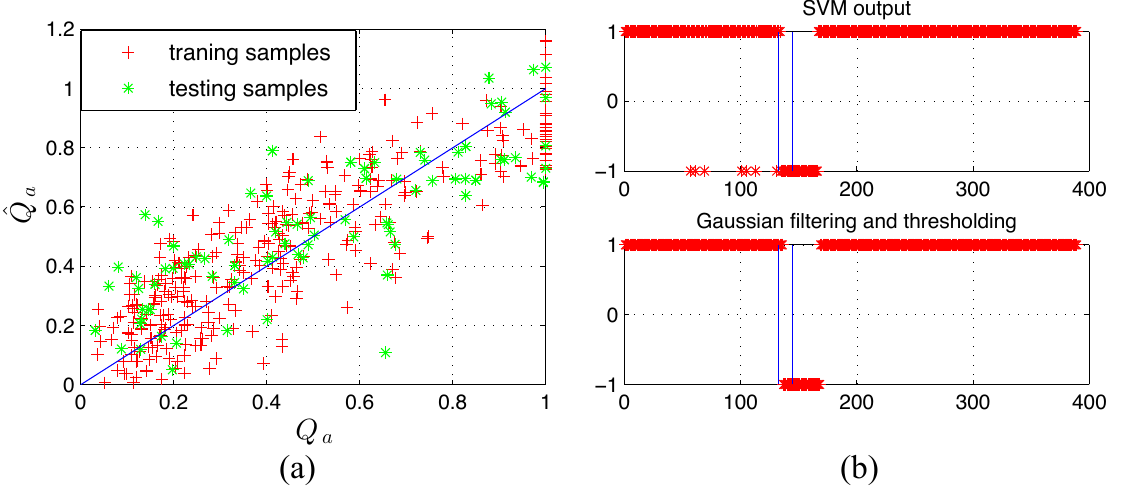}
\end{center}\vspace{-3mm}
\caption{\small (a) Alignment quality model applied to training and testing samples; (b) Tracking quality model applied to one video: the SVM classifier output (top), the result after Gaussian filtering and thresholding (bottom), and the two blue lines are the labeled period of a tracking failure.}
\label{fig:regress}
\end{figure}

\noindent{\bf{Alignment Quality Model}}
Data samples for evaluating our alignment quality model are selected from ${\bm{e}}_1$ in Algorithm~\ref{alg_performance}, which contains the tip-based errors of all leaf pairs with $90\%$ of them are less than $0.5$.
We select $100$ samples from ${\bf{e}}_1$ for each interval of tip-based error within $[0:0.1:0.5]$.
Sample duplication is employed when the number of sample in a particular interval is less than $100$.
All samples with tip-based error larger than $0.5$ will also be selected but without duplication.
Finally we select $625$ samples and extract features $\bm{x}_a$ for each sample.
We assign $Q_a=2\bm{e}_{la}$ to make the output in the range of $[0,1]$.
And $Q_a=1$ for all samples with ${\bm{e}}_{la}>0.5$.
We randomly select $100$ samples as the test set and the remaining samples are used to train the model.
Figure~\ref{fig:regress} (a) shows the results of the model on both training and testing samples.

We use $\bm{R}^2$ to measure how well the model fits our data. 
It is defined as:
\begin{equation}
\bm{R}^2=1-\frac{\sum{(Q_a-\hat{Q}_a)^2}}{\sum{(Q_a-\bar{Q}_a)^2}},
\label{equ:R2}
\end{equation}
where $\hat{Q}_a$ is the predicted quality value and $\bar{Q}_a$ is the mean of $Q_a$.
In our model, $\bm{R}^2=0.769$ and the correlation coefficients for all testing samples is $0.813$.
Both values indicate a high correlation of $Q_a$ and $\hat{Q}_a$.
This quality model is used to predict the alignment accuracy and generate one predicted curve for each leaf, as shown in Figure~\ref{fig:overview}.

\noindent{\bf{Tracking Quality Model}}
We visualize the results of our method and find $15$ videos that have a tracking failure of one leaf.
As the goal for tracking quality model is to detect when the tracking failure starts, we label two frames when the failure starts and ends in each video.
The starting frame is when a leaf candidate starts to change its location toward its neighboring leaves.
The ending frame is when a leaf candidate totally overlaps another leaf.
Among all failure samples, the shortest tracking failure length is $5$ frames and the average length is $12$ frames.

We select $2$-$3$ frames near the ending frame as the negative training samples with $Q_t=-1$ and $5$ frames evenly distributed before the failure starts as the positive training samples with $Q_t=1$.
The features $\bm{x}_t$ are extracted as discussed in Section~\ref{SVM} and used to train a SVM classifier.
The learned model is applied to all frames to predict the tracking quality.
Figure~\ref{fig:regress} (b) shows an example of the output.
We apply a Gaussian filter to remove outliers and delete the failure whose length is less than $5$ frames (the shortest length of failure samples).

We compare the first frame of a predicted failure with that of a labeled failure.
When their distance is less than $12$ frames (the average length of the failure samples), it is considered as a true detection.
Otherwise it is a false detection.
Using the leave-one-video-out testing scheme, the quality model generates $11$ true detections and $16$ false detections over $15$ labeled failures.
Similarly, this quality model is applied during tracking and outputs a prediction curve for each leaf (shown in Figure~\ref{fig:overview}).\\

\subsubsection{Limitation analysis}
\begin{figure}[t!]
\centering
\begin{tabular}{@{}c@{}c@{}c@{}c@{}c@{}c@{}}
\includegraphics[trim=102 110 135 50, clip, width=.077\textwidth]{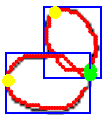} &
\includegraphics[trim=102 110 135 50, clip, width=.077\textwidth]{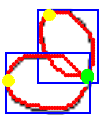} &
\includegraphics[trim=102 110 135 50, clip, width=.077\textwidth]{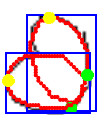} &
\includegraphics[trim=102 110 135 50, clip, width=.077\textwidth]{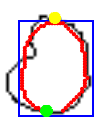} &
\includegraphics[trim=102 108 135 52, clip, width=.077\textwidth]{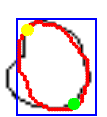} &
\includegraphics[trim=100 108 137 52, clip, width=.077\textwidth]{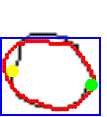} \\
\end{tabular}\vspace{1mm}
\caption{\small Leaf alignment results on synthetic leaves with various amount of overlap. From left to right, the overlap ratio w.r.t.~the smaller leaf is $10\%$, $15\%$, $22\%$, $23\%$, $36\%$, and $59\%$.}
\label{fig:exampleSyn} \figvspace
\end{figure}

Any algorithm has its limitation.
Hence, it is important to explore the limitation of the proposed method.
First, one interesting question is to what extend our segmentation and alignment method can correctly segment leaves in the overlapping region.
We answer this question using a simple synthetic example.
As shown in Figure~\ref{fig:exampleSyn}, our method performs well when the overlap ratio is less than $23\%$.
Otherwise it identifies two leaves as one leaf, which appears to be reasonable when the overlap ratio is high (e.g., $59\%$).

\begin{figure*}[t!]
\begin{center}
\includegraphics[width=.85\textwidth]{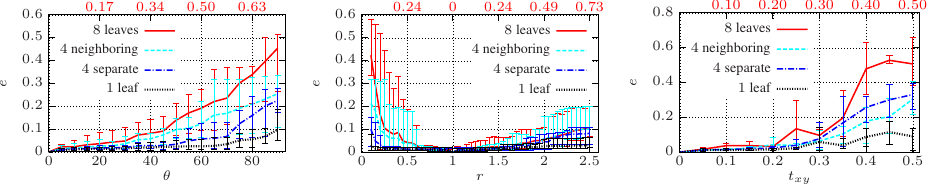}
\end{center} \vspace{-3mm}
\caption{\small Mean tip-based error with different initializations. The axes on top of the figures show the initial tip-based errors.}
\vspace{-3mm}
\label{fig:trackLimit}
\end{figure*}

\begin{figure*}[t!]
\begin{center}
\includegraphics[trim=0 0 0 0, clip, width=.99\textwidth]{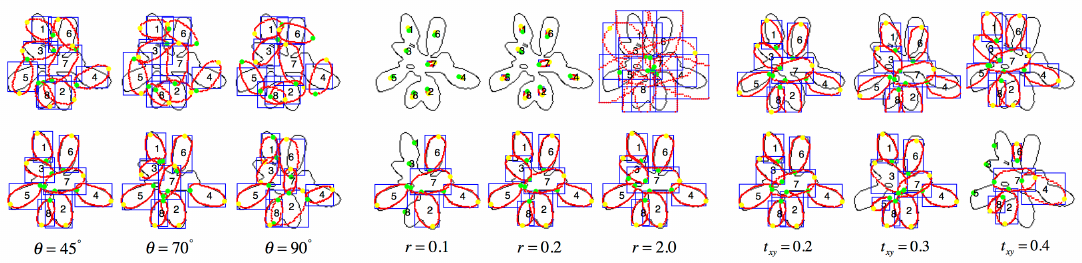}
\end{center} 
\vspace{-3mm}
\caption{\small Example results: the first row shows the initialization, and the second row shows the tracking results.}
\vspace{-2mm}
\label{fig:trackExample}
\end{figure*}

Second, our leaf tracking starts from a good initialization of the leaf candidates from the previous frame.
Another interesting question is to what extend our tracking method can succeed with bad initializations.
To study this, $6$ frames with good tracking results are selected from $6$ videos (one for each).
We change the transformation parameters $\bm{P}$ to synthesize different amount of distortions and apply our tracking algorithm on these $6$ frames.
The leaf candidate is deleted only if it becomes one point and the tip-based error is set to be $1$.
We compute the average tip-based error of all leaf candidates.

We vary the rotation angle $\theta$, the scaling factor $r$, and the translation ratio $t_{xy}$, which is defined as $t_{xy}=\frac{\sqrt{t_x^2+t_y^2}}{\sqrt{(t_1^x-t_2^x)^2+(t_1^y-t_2^y)^2}}$ and the direction is randomly selected. 
The average and range of the tip-based errors for all $6$ frames are shown in Figure~\ref{fig:trackLimit} .
Our tracking method reduces the initial tip-based error to a small value.
It is most robust to $r$ and most sensitive to $t_{xy}$.

Figure~\ref{fig:trackExample} shows some examples.
For rotation angle less than $45^{\circ}$, our method works well for different amounts of leaf rotations.
For the scaling factor, as long as the leaf candidate is not too small, our method is very robust even if we enlarge the original leaf candidates to be $2.5$ times larger.
For the translation ratio, it is sensitive because the direction is randomly selected and leaf candidates are very likely to shift to the locations of the neighboring leaves.
Furthermore, changing the initialization of $\theta$ and $r$ for $4$ separate leaves (leaf $1,4,5,8$ in Figure~\ref{fig:trackExample}) leads to better performance than that of $4$ neighboring leaves (leaf $1,3,6,7$ in Figure~\ref{fig:trackExample}) because neighboring leaves will have overlap with each other and therefore influence the tracking results.
Overall, as the distortion increases, the average tip-based error increases while some of the leaf candidates can still be well aligned.

\section{Conclusions}
In this paper, we identify a new computer vision problem of leaf segmentation, alignment, and tracking from fluorescence plant videos.
Leaf alignment and tracking are formulated as two optimization problems based on Chamfer matching and leaf template transformation.
Two models are learned to predict the quality of leaf alignment and tracking.
A quantitative evaluation scheme is designed to evaluate the performance.
The limitations of our algorithm are studied and experimental results show the effectiveness, efficiency, and robustness of the proposed method.

With the leaf boundary and structure information over time, the {\it photosynthetic efficiency} can be computed for each leaf, which paves the way for leaf-level photosynthetic analysis and high-throughput plant phenotyping.
The proposed method and the evaluation scheme are potentially applicable to other plant videos, as shown in the results on the LSC dataset.

%%%

{\small
\bibliographystyle{IEEEbib}
\bibliography{abbrev,egbib}
}

\vspace{-14mm}
\begin{IEEEbiography}[{\includegraphics[width=1in,height=1.25in,clip,keepaspectratio]{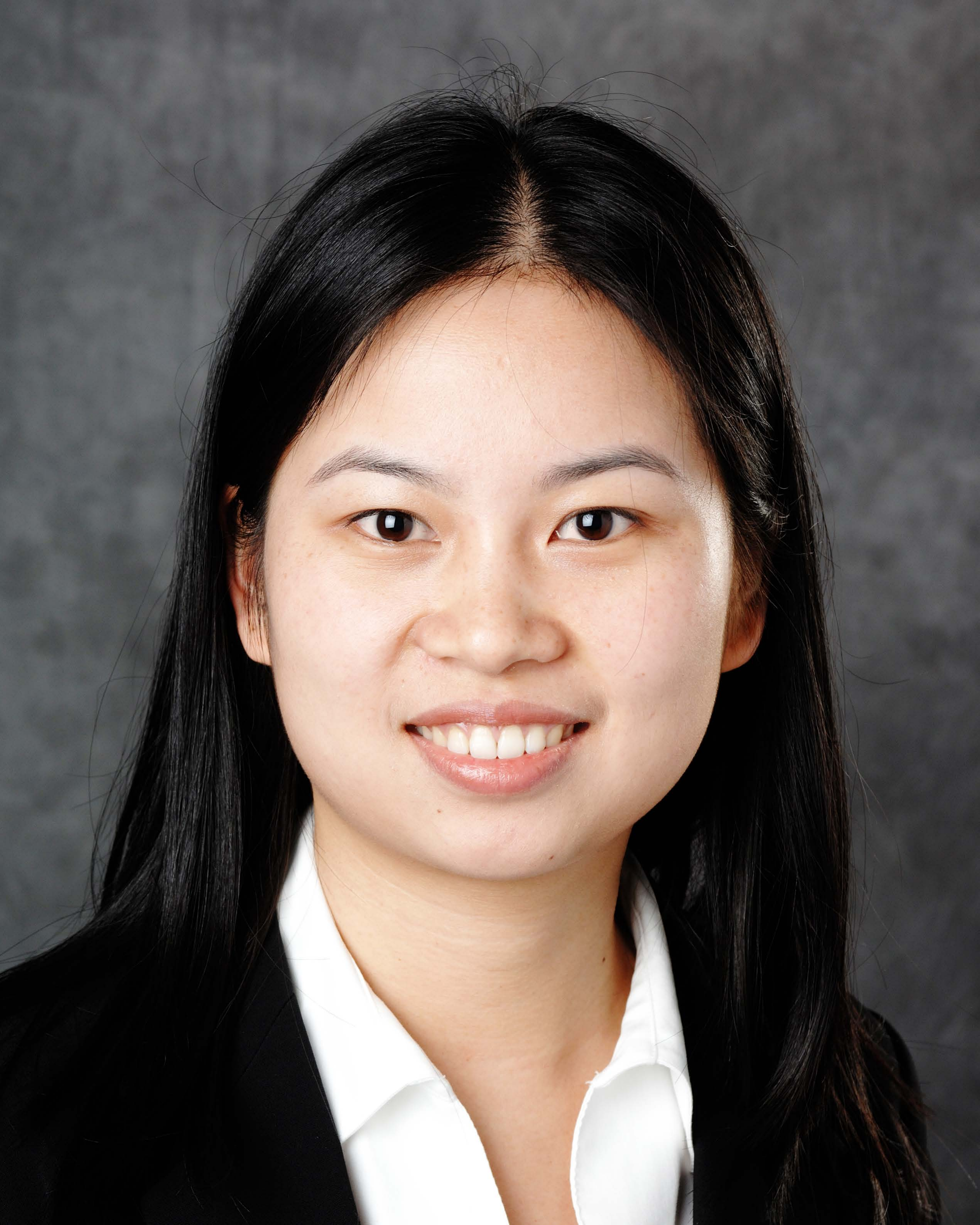}}]
{Xi Yin} received the B.S. degree in Electronic and Information Science from Wuhan University, China, in 2013. Since August 2013, she has been working toward her Ph.D. degree in the Department of Computer Science and Engineering, Michigan State University, USA. Her paper on plant segmentation won the Best Student Paper Award at Winter Conference on Application of Computer Vision (WACV) 2014. 
Her research interests include computer vision and deep learning.
\end{IEEEbiography}

\vspace{-12mm}
\begin{IEEEbiography}[{\includegraphics[width=1in,height=1.25in,clip,keepaspectratio]{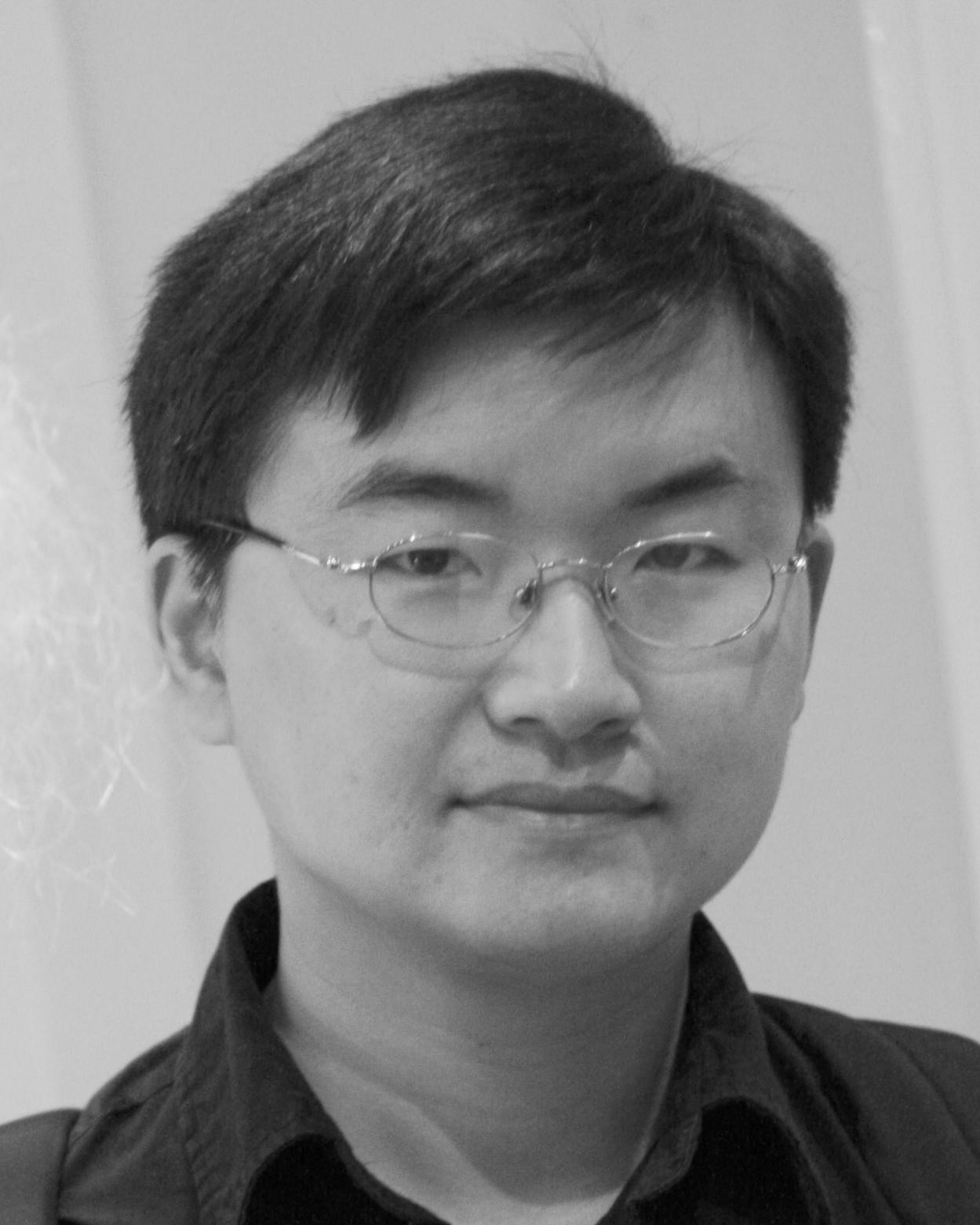}}]
{Xiaoming Liu} is an Assistant Professor in the Department of Computer Science and Engineering at Michigan State University (MSU). He received the B.E. degree from Beijing Information Technology Institute, China and the M.E. degree from Zhejiang University, China, in $1997$ and $2000$ respectively,
both in Computer Science, and the Ph.D. degree in Electrical and Computer Engineering from Carnegie Mellon University in $2004$. Before joining MSU in Fall $2012$, he was a research scientist at General Electric Global Research Center. His research areas are face recognition, biometrics, image alignment, video surveillance, computer vision and pattern recognition. He has authored more than $70$ scientific publications, and has filed $22$ U.S. patents. He is a member of the IEEE.
\end{IEEEbiography}

\vspace{-12mm}
\begin{IEEEbiography}[{\includegraphics[width=1in,height=1.25in,clip,keepaspectratio]{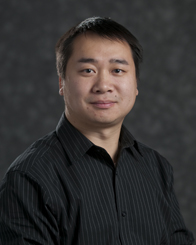}}]
{Jin Chen} received the BS degree in computer science from Southeast University, China, in $1997$, and the Ph.D. degree in computer science from the National University of Singapore, Singapore, in $2007$. He is an Assistant Professor in the Department of Energy Plant Research Laboratory and the Department of Computer Science and Engineering at Michigan State University. His general research interests are in computational biology, as well as its interface with data mining and computer vision.
\end{IEEEbiography}

\vspace{-12mm}
\begin{IEEEbiography}[{\includegraphics[width=1in,height=1.25in,clip,keepaspectratio]{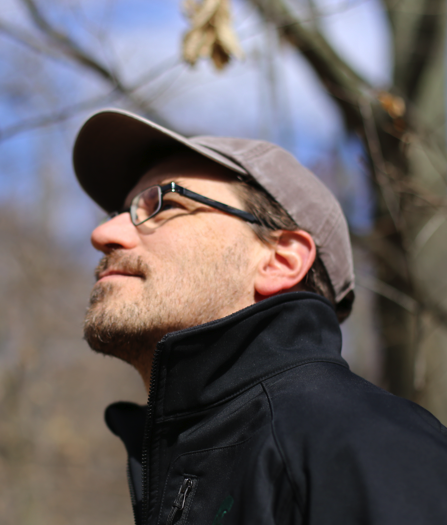}}]
{David M. Kramer} is the Hannah Distinguished Professor of Bioenergetics and Photosynthesis in the Biochemistry and Molecular Biology Department and the MSU-Department of Energy Plant Research Lab at Michigan State University. In 1990, he received his Ph.D. in Biophysics at University of Illinois, Urbana-Champaign, followed by Post-doc at the Institute de Biologie Physico-Chimique in Paris and a 15-year tenure as a faculty member at Washington State University. His research seeks to understand how plants convert light energy into forms usable for life, how these processes function at both molecular and physiological levels, how they are regulated and controlled, how they define the energy budget of plants and the ecosystem and how they have adapted through evolution to support life in extreme environments. This work has led his research team to develop a series of novel spectroscopic tools for probing photosynthetic reactions in vivo.
\end{IEEEbiography}

\ifCLASSOPTIONcaptionsoff
\newpage
\fi

\end{document}